\documentclass{article}

    \usepackage[preprint]{neurips_2019}

\usepackage[utf8]{inputenc} 
\usepackage[T1]{fontenc}    
\usepackage[final]{hyperref}       
\usepackage{url}            
\usepackage{booktabs}       
\usepackage{amsfonts}       
\usepackage{nicefrac}       
\usepackage{microtype}      

\usepackage{algorithm,algorithmic}
\usepackage{graphicx}
\usepackage{multirow}
\usepackage{amsmath, bm}
\usepackage{wrapfig}
\usepackage{xfrac}
\usepackage{sidecap}
\usepackage{adjustbox}
\usepackage{caption}
\usepackage{subcaption}

\usepackage[dvipsnames]{xcolor}

\newcommand{\mean}{\mathbb{E}}
\newcommand{\var}{{\rm I\kern-.3em D}}
\newcommand{\Var}{\mathrm{Var}}

\newcommand{\cond}{\,|\,}
\newcommand{\Normal}{\mathcal{N}}

\DeclareMathOperator*{\argmin}{arg\,min}
\newtheorem{theorem}{Theorem}

\title{Metropolis-Hastings View \\ on Variational Inference and Adversarial Training}

\author{%
Kirill Neklyudov \\
Samsung AI Center Moscow \\
Samsung-HSE Laboratory \\
HSE\thanks{National Research University Higher School of Economics}, \; Moscow, Russia \\
\texttt{k.necludov@gmail.com}
  \And
Evgenii Egorov \\
Skoltech\thanks{Skolkovo Institute of Science and Technology}, \; Moscow, Russia \\
\texttt{egorov.evgenyy@ya.ru}
  \And
Dmitry Vetrov \\
Samsung AI Center Moscow \\
Samsung-HSE Laboratory\\
HSE\footnotemark[1], \; Moscow, Russia \\
\texttt{vetrovd@yandex.ru}
}

\begin{document}

\maketitle

\begin{abstract}

A significant part of MCMC methods can be considered as the Metropolis-Hastings (MH) algorithm with different proposal distributions.
From this point of view, the problem of constructing a sampler can be reduced to the question --- how to choose a proposal for the MH algorithm?
To address this question, we propose to learn an independent sampler that maximizes the acceptance rate of the MH algorithm, which, as we demonstrate, is highly related to the conventional variational inference.
For Bayesian inference, the proposed method compares favorably against alternatives to sample from the posterior distribution.
Under the same approach, we step beyond the scope of classical MCMC methods and deduce the Generative Adversarial Networks (GANs) framework from scratch, treating the generator as the proposal and the discriminator as the acceptance test.
On real-world datasets, we improve Frechet Inception Distance and Inception Score, using different GANs as a proposal distribution for the MH algorithm.
In particular, we demonstrate improvements of recently proposed BigGAN model on ImageNet.

\end{abstract}

\section{Introduction}
The problem of sampling from a distribution is one of the key tasks in building models from data.
Sampling techniques find direct application in simulation of physical systems such as spin glasses \citep{ogielski1985dynamics} and protein folding problems \citep{mitsutake2001generalized}.
They serve as building blocks of machine learning algorithms, especially in Bayesian inference \citep{mackay2003information}.
Finally, learning a generative model from a distribution that is given in the empirical form (as a set of samples) is a way to describe this distribution, generate new samples \citep{goodfellow2014generative} and learn representations \citep{kingma2013auto, donahue2016adversarial}.

To sample from target distributions that are only analytically tractable up to a normalizing constant one usually refers to Markov Chain Monte Carlo (MCMC) methods.
They operate by generating a chain of correlated samples that converge in distribution to the target.
This convergence is most often guaranteed through the detailed balance condition, a sufficient condition for the chain to have the target equilibrium distribution.
In practice, the Metropolis-Hastings (MH) correction \citep{hastings1970monte} is usually employed to ensure the detailed balance condition for a broad class of proposal distributions.
The price we pay for obtaining convergence in such a general way is that we need to choose the proposal wisely to obtain an efficient sampler.
Speaking informally, there are two main issues while choosing a proposal: the proposal should generate diverse samples to explore the target distribution quickly; these samples should be accepted frequently to meet the available computational budget.

In this paper, we achieve the diversity of samples by considering independent proposal distributions that generate uncorrelated samples.
Although this restriction may seem too limiting, there are plenty of successful illustrations of probabilistic models parameterized by deep neural networks that sample independently from highly complex and multi-modal distributions \citep{goodfellow2014generative, kingma2013auto, rezende2015variational, dinh2016density}.

Having an independent proposal, we propose to maximize the acceptance rate of the MH algorithm directly, w.r.t. parameters of a neural network that simulates a proposal distribution.
Moreover, we prove that the acceptance rate can be lower bounded via negative symmetrized KL divergence between proposal and target.
We demonstrate empirically that learning a proposal by minimization of symmetrized KL divergence compares favorably against minimization of reversed KL divergence (the variational inference) as was proposed in \cite{de2001variational}.
Intuitively, this makes sense: reverse KL divergence is known to be mode-seeking, which could greatly hurt the performance of the MH algorithm.
To prevent such behavior, we should take into account forward KL divergence that fosters mass covering.

Starting from the same idea of learning a proposal for the MH algorithm, we derive GANs framework \citep{goodfellow2014generative} from scratch.
Recall that the GANs framework assumes target distribution to be given as a set of samples, rather than as unnormalized density.
It turns out that treating GANs from the MCMC perspective by itself improves their performance.
In particular, we demonstrate that taking the generator network as a proposal distribution and running the MH algorithm via the learned discriminator, improves quality of samples in terms of Frechet Inception Distance (FID) \citep{heusel2017gans} and Inception Score (IS) \citep{salimans2016improved}.

The main contributions of our paper can be summarized as follows.
\begin{itemize}
    \item We prove that the acceptance rate of the MH algorithm with an independent proposal can be lower bounded using symmetrized KL-divergence between proposal and target (Section \ref{sec:ARMH}).
    \item We propose an algorithm based on the maximization of the acceptance rate for learning a proposal distribution of the MH algorithm (Section \ref{sec:db_algs}).
    We deduce the GANs framework from scratch, using the generator as the proposal and the discriminator for the acceptance test (Section \ref{sec:sb_algs}).
    \item For the proposed approach, we demonstrate empirical gains while sampling from the posterior distribution of Bayesian logistic regression (Section \ref{sec:exps_db}).
    Applying the MCMC formalism to GANs, we improve FID and IS for different GANs on real-world datasets (Section \ref{sec:exps_sb}).
    For large scale validation, we demonstrate empirical gains on the recently developed BigGAN \citep{brock2018large}.
\end{itemize}

\section{Related work}
\label{sec:RW}

Since the choice of the proposal is the crux of the MH algorithm efficiency, many works have addressed this issue in various ways.
Classical works of \cite{roberts1997weak,roberts2001optimal} suggest a general guideline for scaling the variance of a random-walk proposal with the growth of target dimensionality.
More complex proposal designs include adaptive updates of the proposal distribution during iterations of the MH algorithm, what is known as the adaptive MH algorithm \citep{holden2009adaptive,giordani2010adaptive}.
Another way to adapt the MH algorithm for complex distributions is a combination of adaptive direction sampling and the multiple-try Metropolis algorithm as proposed in \citep{liu2000multiple}.
Thorough overview of different extensions of the MH algorithm is presented in \citep{martino2018review}.
Among works on the proposal choice, the closest to our is Variational MCMC \citep{de2001variational}.
They suggest to learn an independent proposal using the variational inference procedure and then enrich the learned proposal by mixing it with a random walk kernel.
Compared to these works, we consider deep neural networks as a proposal distribution and propose a new algorithm to train them.

Several works have been done in a similar direction but to learn a transition kernel in Hybrid Monte-Carlo (HMC) \citep{duane1987hybrid}. 
A-NICE-MC algorithm \citep{song2017nice} is HMC-inspired method that uses the family of NICE networks \citep{dinh2014nice} as transition kernels that authors propose to learn via adversarial training.
\cite{levy2017generalizing} propose L2HMC algorithm that achieves flexible transition kernels by incorporating neural networks directly into the leap-frog integrator.
These algorithms outperform the vanilla HMC algorithm; hence, we consider these works as baselines for the performance evaluation of our model.

Although MCMC and GAN are orthogonal approaches, they share the common goal --- learning to sample.
This fact, motivates the work of \citep{song2017nice}, where they improve an MCMC sampler via adversarial training.
It also inspires the application of MCMC techniques for improvement of GAN \cite{azadi2018discriminator}.
In this work, authors propose to use the rejection sampling algorithm to improve the performance of a GAN.
However, applying rejection sampling to a GAN raise the issue of finding the majorization constant, that cannot be evaluated efficiently in practice.
It is also clear that the MH algorithm has a higher acceptance rate for the same proposal distribution.

\section{Background}
\label{sec:back}
\textbf{The MH algorithm} generates a chain of samples that converges to an analytic target distribution $p(x)$ while one is only able to sample from a proposal distribution $q(x'\cond x)$. 
One step of the chain generation can be described as follows. 
\begin{enumerate}
    \item sample proposal point $x' \sim q(x'\cond x)$, given previous point $x$
    \item accept $
    \begin{cases}
    x' , \;\;\; \text{with probability} \;\;\; \min\bigg\{1, \frac{p(x')q(x\cond x')}{p(x)q(x'\cond x)}\bigg\} \\
    x , \;\;\; \text{otherwise}
    \end{cases}$
\end{enumerate}

The computational efficiency of the MH algorithm significantly depends on the probabilities of acceptance.
The averaged probability of acceptance is called the acceptance rate and is defined as 
\begin{align}
    \mathrm{AR} =  \int dxdx' p(x)q(x'\cond x) \min\bigg\{1, \frac{p(x')q(x\cond x')}{p(x)q(x'\cond x)}\bigg\}.
    \label{eq:classic_AR}
\end{align}

\textbf{Independent MH} is an important case of the MH algorithm.
The only difference is that independent MH does not condition a proposal point $x'$ on the previous point $x$, i.e. $q(x' \cond x) = q(x')$.
An attractive property of independent proposals is their ability to make large jumps, and if this can be done while keeping the acceptance rate high, the autocorrelation of the chain will be small.

\textbf{Rejection sampling} can be considered as an alternative to independent MH.
For proposal distribution $q(x)$ such that $\exists M \in \mathbb{R}: q(x)M \geq p(x) \; \forall x$, rejection sampling generates the next sample in the following way
\begin{enumerate}
    \item sample proposal point $x' \sim q(x')$
    \item accept new point $x'$ with probability $\frac{p(x')}{Mq(x')}$
\end{enumerate}

It is worth mentioning that rejection sampling, and independent MH are directly comparable in terms of the acceptance rate. The average number of accepted points in rejection sampling is
\begin{align}
    \mathrm{AR}_{\mathrm{RS}} = \int dx' q(x') \frac{p(x')}{Mq(x')} = \frac{1}{M}.
\end{align}
At the same time, for independent MH with the same proposal $q$, the average number of accepted points can be lower bounded as
\begin{equation}
    \mathrm{AR}_{\mathrm{IMH}} = \int dxdx' p(x)p(x') \min\bigg\{\frac{q(x')}{p(x')}, \frac{q(x)}{p(x)}\bigg\} \geq \frac{1}{M} = \mathrm{AR}_{\mathrm{RS}}.
\end{equation}
Together with the difficulties of finding appropriate $M$, this fact motivates the usage of independent MH instead of rejection sampling.

\section{The lower bound on the acceptance rate}
\label{sec:ARMH}

The acceptance rate of the MH algorithm is tightly connected with detailed balance.
In the extreme case when the acceptance rate achieves its maximum value, distributions $p(x')q(x\cond x')$ and $p(x)q(x'\cond x)$ must coincide (up to sets of zero measure) in the joint space of the previous point $x$ and the proposed point $x'$.
For such a case, we can say that the detailed balance condition holds:
\begin{equation}
    p(x')q(x\cond x') = p(x)q(x'\cond x) \;\;\; \forall x,x'.
\end{equation}
It turns out that the acceptance rate defines how far distributions $p(x')q(x\cond x')$ and $p(x)q(x'\cond x)$ are, or how well the detailed balance condition is satisfied for a proposal distribution $q(\cdot\cond\cdot)$.
We formalize this connection by introducing the following theorem.
\begin{theorem}
For a random variable $\xi = \frac{p(x')q(x\cond x')}{p(x)q(x'\cond x)}, x \sim p(x), x' \sim q(x'\cond x)$
\begin{align}
    \mathrm{AR}=\mean_\xi \min\{1, \xi\} = 1 - \frac{1}{2} \mean_\xi |\xi - 1| = 1 - \mathrm{TV}\bigg(p(x')q(x\cond x') \bigg\Vert p(x)q(x'\cond x)\bigg),
\end{align}
where $\mathrm{TV}$ is the total variation distance.
\label{th}
\end{theorem}
See proof of Theorem \ref{th} in Appendix \ref{app:proof}. 
This reinterpretation in terms of total variation allows us to lower bound the acceptance rate via Pinsker's inequality
\begin{equation}
    \mathrm{AR} \geq 1 - \sqrt{\frac{1}{2} \cdot \text{KL}\bigg(p(x')q(x\cond x') \bigg\Vert p(x)q(x'\cond x)
    \bigg)}.
\label{eq:arlb}
\end{equation}
We suggest using the acceptance rate or its lower bound as an objective for learning a proposal distribution.
Note that doing so for a Markov proposal may result in a trivial solution $q(x'\cond x) = \delta(x'-x)$ that yields the maximal acceptance rate.
That happens, since detailed balance and the acceptance rate does not take the autocorrelation of proposed samples into account.
In this work, we enforce zero autocorrelation and exclude the trivial solution by considering independent proposals.
For more details, see Appendix \ref{app:independent}.

For independent proposals the lower bound from Eq. \ref{eq:arlb} can be rewritten in terms of symmetric KL-divergence between $p(\cdot)$ and $q(\cdot)$
\begin{equation}
    \mathrm{AR} \geq 1 - \sqrt{\frac{1}{2}\bigg(\text{KL}(q(x)\Vert p(x)) + \text{KL}(p(x)\Vert q(x))\bigg)},
\label{eq:arlb_ind}
\end{equation}
which has its global maximum at $q(x)=p(x)$; hence, at maximal acceptance rate $\mathrm{AR} = 1$.
In Section \ref{sec:algs}, we demonstrate that the obtained lower bound relates the proposed approach with the variational inference and GANs.
For Bayesian inference, the obtained lower bound could be preferable to the acceptance rate since one can estimate it using only minibatches of data.

Additionally, we show that in the case of an independent proposal, the acceptance rate defines a semimetric in distribution space between $p(\cdot)$ and $q(\cdot)$ (see Appendix \ref{app:semi}).

\section{Optimization of proposal distribution}
\label{sec:algs}

In this section, we propose algorithms for learning parameters $\phi$ of an independent proposal distribution $q_\phi(x)$.
As objectives for optimization, we use the acceptance rate of the MH algorithm and its lower bound.
For convenience, we reformulate these objectives in terms of loss functions as follows.
Maximization of the acceptance rate is equivalent to minimization of the loss:
\begin{align}
    \mathcal{L}_{\mathrm{AR}}(\phi) = -\mathrm{AR} = -\text{\Large $\mean$}_{\text{\footnotesize $\begin{matrix}
    x \sim p(x) \\ x' \sim q_\phi(x')\end{matrix}$}} \min\bigg\{1, \frac{p(x')q_\phi(x)}{q_\phi(x')p(x)}\bigg\}.
    \label{eq:loss_ar}
\end{align}
For maximization of the lower bound, the loss function is
\begin{align}
    \mathcal{L}_{\mathrm{KL}}(\phi) = \text{KL}(q_\phi(x)\Vert p(x)) + \text{KL}(p(x)\Vert q_\phi(x)) = -\text{\Large $\mean$}_{\text{\footnotesize $\begin{matrix}
    x \sim p(x) \\ x' \sim q_\phi(x')\end{matrix}$}} \log\bigg( \frac{p(x')q_\phi(x)}{q_\phi(x')p(x)}\bigg).
    \label{eq:loss_lb}
\end{align}
We will refer to both of these losses as $\mathcal{L}(\phi)$, assuming that any of them can be substituted.
We also denote $l(\cdot)$ as any of $\min\{1,\cdot\}$ and $\log(\cdot)$.

To estimate $\mathcal{L}(\phi)$, we need to evaluate the density ratio on samples from the target $x \sim p(x)$ and proposal $x' \sim q_\phi(x')$.
Depending on the form in which the target distribution is given, we have different issues during the estimation of the loss function.

If the target distribution is given as an unnormalized density (we call it \textbf{the density-based setting}), we suggest using an explicit probabilistic model as a proposal to evaluate the density ratio exactly. 
To obtain samples from the target, in this setting we propose to run independent MH with the currently available proposal. 

If the target distribution is given in the empirical form (we call it \textbf{the sample-based setting}), samples from the target and proposal distributions are available, but we cannot compute the density ratio, so we propose to approximate it via the adversarial training.

For a summary of both settings, see Table \ref{tab:settings}.
The following subsections describe algorithms in detail.

\begin{table*}[t]
\def\arraystretch{1.5}
\caption{Short discription of two settings for target distribution $p(x)$ and proposal $q(x)$.}
\label{tab:settings}
\begin{center}
\vskip -0.05in
\resizebox{\textwidth}{!}{
\begin{tabular}{l|llll}
\toprule
{\bf Setting} & {\bf Density of target} &{\bf Samples from target} & {\bf Density of proposal} & {\bf Density ratio}  \\
\midrule
Density-based & given          & run independent MH & given         & given \\ \hline
Sample-based  & not available  & given  & not available & run adversarial training \\
\bottomrule
\end{tabular}}
\end{center}
\end{table*}

\subsection{Density-based setting}
\label{sec:db_algs}

In the density-based setting, we assume the proposal to be an explicit probabilistic model, i.e. the model that we can sample from and evaluate its density at any point up to the normalization constant.
We also assume that the proposal is reparameterizable \citep{rezende2014stochastic}.


\begin{wrapfigure}{R}{0.5\textwidth}
    \begin{minipage}{0.5\textwidth}
    \vskip -0.3in
      \begin{algorithm}[H]
      \caption{Learning the proposal distribution in the density-based setting}
      \begin{algorithmic}  
        \INPUT{density of target distribution $\hat{p}(x) \propto p(x)$}
        \INPUT{explicit probabilistic model $q_\phi(x')$}
        \STATE Initialize \textit{Buffer}
        \FOR{$n$ iterations}
            \STATE add new samples to \textit{Buffer} 
            \STATE \COMMENT{using the MH with the current proposal $q_\phi$}
            \STATE sample $\{x_k\}_{k=1}^K \sim p(x)$ from \textit{Buffer}
            \STATE sample $\{x'_k\}_{k=1}^K \sim q_\phi(x')$ 
            \STATE $\mathcal{L}(\phi) \simeq - \frac{1}{K}\sum_{k=1}^K l\bigg( \frac{p(x'_k)q_\phi(x_k)}{p(x_k)q_\phi(x'_k)}\bigg)$
            \STATE $\phi \gets \phi - \alpha \nabla_\phi \mathcal{L}(\phi)$
        \ENDFOR
        \OUTPUT{parameters $\phi$}
      \end{algorithmic} 
      \label{alg:db}
    \end{algorithm}
    \end{minipage}
\end{wrapfigure}

We consider normalizing flows \citep{rezende2015variational, dinh2016density} as explicit proposal distributions.
The rich family of normalizing flows allows us to learn an expressive proposal and evaluate its density at any point in the target distribution space.
Taking invertible models as proposals, we guarantee the ergodicity of the resulting Markov chain, since they cover all of the target space with positive density.
Indeed, choosing an arbitrary point in the target space, we can obtain the corresponding point in the latent space using the inverse function.
If we sample from a Gaussian in a latent space, then every point in the target space has positive density.

Explicit proposal $q_\phi(x')$ and target $p(x)$ distributions allow for the accurate density ratio evaluation but to estimate loss $\mathcal{L}(\phi)$, we also need samples from the target. 
For this purpose, we use the currently learned proposal $q_\phi$ and collect samples via the MH algorithm. 
We aggregate samples in a buffer throughout the learning, that allows us to cut the computational budget.
After obtaining samples from the target distribution, it is possible to perform the optimization step by taking stochastic gradients w.r.t.~$\phi$.
We provide a pseudo-code in Algorithm \ref{alg:db}.

Now we apply this algorithm for the Bayesian inference and show that during optimization of the lower bound \eqref{eq:loss_lb} we can use only minibatches of data, while it is not the case for the direct optimization of acceptance rate.
Given a probabilistic model $p(y\cond x, \theta)$, we want to tune parameters $\theta$ on a dataset $\mathcal{D} = \{(x_i, y_i)\}_{i=1}^N$. 
The distribution $p(\theta)$ depicts prior knowledge about $\theta$. 
Following the Bayesian approach, we first need to infer the posterior distribution $p(\theta\cond \mathcal{D})$ and then obtain predictive distribution as
\begin{equation}
    p(y\cond x) = \mean_{p(\theta\cond \mathcal{D})} p(y\cond x, \theta).
    \label{eq:pred_dist}
\end{equation}
Since the posterior $p(\theta\cond \mathcal{D})$ is intractable in most cases, one usually resorts to the Monte-Carlo estimation of the formula \eqref{eq:pred_dist}.
To sample from the posterior, we suggest using the MH algorithm, learning a proposal $q_\phi(\theta)$ by the maximization of acceptance rate lower bound \eqref{eq:arlb_ind}.
Objective \eqref{eq:loss_lb} for this problem becomes
\begin{align}
    \mathcal{L}_{\mathrm{KL}}(\phi) = \text{KL}\bigg(q_\phi(\theta) \bigg\Vert p(\theta\cond \mathcal{D})\bigg) + \text{KL}\bigg(p(\theta'\cond \mathcal{D})\bigg\Vert q_\phi(\theta')\bigg)
\end{align}
Minimization of the reversed KL divergence (the first term) corresponds to the variational inference and results in a mode-seeking solution.
To maximize the acceptance rate, we need to add the forward KL divergence (the second term) that fosters mass covering.
Getting rid of the terms that do not depend on $\phi$, we obtain the equivalent optimization problem
\begin{align}
    \min_\phi\bigg[&-\mean_{\theta \sim q_\phi(\theta)} \sum_{i=1}^N \log p(y_i\cond x_i,\theta) + \text{KL}(q_\phi(\theta) \Vert p(\theta)) - \mean_{\theta \sim p(\theta\cond \mathcal{D})} \log q_\phi(\theta)\bigg].
    \label{eq:db_inference}
\end{align}
This objective allows for the unbiased minibatch estimation.
Indeed, one can estimate the first two terms by following the doubly stochastic variational inference \citep{titsias2014doubly}.
The estimation of the last term seems to be more challenging since it relates to the samples from the posterior.
Fortunately, there are minibatch versions of the Metropolis-Hastings algorithm \citep{korattikara2014austerity, chen2016efficient}, that one can use to obtain samples from the posterior.
Combination of these techniques allows us to use only minibatches of data during iterations of Algorithm \ref{alg:db}.

\subsection{Sample-based setting}
\label{sec:sb_algs}

\begin{wrapfigure}{R}{0.5\textwidth}
    \begin{minipage}{0.5\textwidth}
    \vskip -0.3in
      \begin{algorithm}[H]
      \caption{Learning the proposal distribution in the sample-based setting}  
      \begin{algorithmic}  
        \INPUT{set of samples $X \sim p(x)$}
        \INPUT{implicit probabilistic model $q_\phi(x)$}
        \FOR{$n$ iterations}
            \STATE sample $\{x_k\}_{k=1}^K \sim X$
            \STATE sample $\{x'_k\}_{k=1}^K \sim q_\phi(x')$ 
            \STATE train the discriminator $d$ as in \ref{eq:disc_opt}
            \STATE $\mathcal{L}(\phi) \approx - \frac{1}{K}\sum_{k=1}^K l\bigg( \frac{d(x_k')(1-d(x_k))}{(1-d(x_k'))d(x_k)}\bigg)$ 
            \STATE $\phi \gets \phi - \alpha \nabla_\phi \mathcal{L}(p, q_\phi)$
        \ENDFOR\\
        \OUTPUT parameters $\phi$
      \end{algorithmic} 
      \label{alg:sb}
    \end{algorithm}
    \end{minipage}
\end{wrapfigure}
In the sample-based setting, we demonstrate that the GAN framework can be derived from scratch using the MCMC perspective.
As well as in GAN, we assume the proposal $q_\phi(x)$ to be an implicit probabilistic model (we have access to the samples but not to the density) and the target distribution $p(x)$ is given in the empirical form (as a dataset).

Under the assumptions mentioned above, we have samples both from $q_\phi(x)$ and $p(x)$ but cannot evaluate the density ratio $p(x)/q_\phi(x)$.
However, it is still possible to estimate the density ratio by learning a network $d(x)$, that discriminates between samples from target and proposal.
The optimal discriminator $d^*(x)$ yields
\begin{align}
    d^*(x) = \frac{p(x)}{p(x)+q(x)} = \argmin_d \bigg[-\mean_{x\sim p(x)} \log d(x) -\mean_{x \sim q(x)} \log (1-d(x)) \bigg].
    \label{eq:disc_opt}
\end{align}
Using this equation, we approximate the acceptance ratio as
\begin{equation}
    \frac{p(x_k')q(x_k)}{q(x_k')p(x_k)} \approx \frac{d(x_k')(1-d(x_k))}{(1-d(x_k'))d(x_k)}.
    \label{eq:approx_test}
\end{equation}
We formulate Algorithm \ref{alg:sb} using this approximation for the estimation of $\mathcal{L}(\phi)$.
If we take the loss function $\mathcal{L}_{\mathrm{KL}}(\phi)$ as the objective for the proposal, we obtain
\begin{align}
    \mathcal{L}_{\mathrm{KL}}(\phi) \approx \mean_{x \sim q_\phi(x)} \log (1-d(x)) -\mean_{x\sim q_\phi(x)} \log d(x).
\label{eq:gan_lb}
\end{align}
The first term here is the "zero-sum" loss for the generator in the minimax game, while the second term is the well-known rule of thumb that prevents gradient saturation \citep{goodfellow2016nips}.
If we assume that the discriminator distinguishes the fake samples confidently ($d(x) \approx 0$ for $x\sim q_\phi(x)$), then we end up with a conventional GAN formulation.

This connection motivates us to rethink the sampling from a GAN on the test stage.
Indeed, the generator training resembles the optimization of the proposal distribution in the MH; however, on the test stage, one usually samples using only this proposal.
The right thing to do is to sample via the MH algorithm.
To run the MH algorithm, we propose using the already learned discriminator for the approximate estimation of the acceptance test \eqref{eq:approx_test}.
In Section \ref{sec:exps}, we show that sampling via the MH algorithm demonstrates consistent improvements compared to the sampling from a generator.

We additionally explore the way of learning a Markov proposal in this algorithm but leave a rigorous study of it for future work (see Appendices \ref{app:markov_gan},\ref{app:mnist_gan}).

\section{Experiments}
\label{sec:exps}

We present an empirical evaluation for both density-based and sample-based settings.
In the density-based setting, the proposed algorithm compares favorably in sampling from the posterior distribution of the Bayesian logistic regression.
In the sample-based setting, we demonstrate empirical gains of various GANs by sampling via the MH algorithm at the test stage.
Code reproducing all experiments is available online\footnote{\url{https://github.com/necludov/MH-AR}}.

\subsection{Density-based setting}
\label{sec:exps_db}

\begin{table}[t]
\caption{Comparison with A-NICE-MC and L2HMC on synthetic distributions and the posterior distribution of Bayesian logistic regression. Performance of algorithms as measured by Effective Sample Size (ESS) (1k samples for synthetic, 5k for the posteriors). For computational efforts, we provide ESS per second. Higher values of ESS and ESS per second are better (for the detailed formulation, see Appendix \ref{app:ess}). See the description of the compared models in the text.}
\label{tab:results_db}
\vskip -0.25in
\begin{center}
\begin{small}
\resizebox{\textwidth}{!}{
\begin{tabular}{lcccc|cccc}
    \toprule
    \multicolumn{1}{c}{} & \multicolumn{4}{c}{\bf ESS} & \multicolumn{4}{c}{\bf ESS per second}\\
    \midrule
    {Target}&{A-NICE-MC\footnotemark}&{AR}(ours)&{ARLB}(ours)&{VI}&{A-NICE-MC\footnotemark[\value{footnote}]}&{AR}(ours)&{ARLB}(ours)&{VI}\\
    \midrule
    Ring & $\mathbf{1000}$ & $863$ & $811$ & $717$ & $\mathbf{12.7\cdot 10^5}$ & $5.0\cdot 10^5$ & $4.7\cdot 10^5$ & $4.2\cdot 10^5$ \\ 
    Mog2 & $355$ & $732$ & $\mathbf{746}$ & $297$ & $4.1\cdot 10^5$ & $4.1\cdot 10^5$ & $\mathbf{4.2\cdot 10^5}$ & $1.7\cdot 10^5$ \\
    Mog6 & $320$ & $\mathbf{510}$ & $401$ & $12$ & $\mathbf{2.7\cdot 10^5}$ & $2.4\cdot 10^5$ & $1.9\cdot 10^5$ & $0.06\cdot 10^5$ \\
    Ring5 & $156$ & $\mathbf{336}$ & $249$ & $170$ & $1.5\cdot 10^5$ & $\mathbf{1.7\cdot 10^5}$ & $1.3\cdot 10^5$ & $0.9\cdot 10^5$ \\
    \midrule
    {Dataset}&{A-NICE-MC\footnotemark[\value{footnote}]}&{AR}(ours)&{ARLB}(ours)&{VI}&{A-NICE-MC\footnotemark[\value{footnote}]}&{AR}(ours)&{ARLB}(ours)&{VI}\\
    \midrule
    German & $926$ & $\mathbf{5000}$ & $4105$ & $3535$ & $0.8 \cdot 10^6$ & $\mathbf{2.6 \cdot 10^6}$ & $2.1 \cdot 10^6$ & $1.8 \cdot 10^6$ \\
    Heart & $1251$ & $\mathbf{5000}$ & $4362$ & $4110$ & $1.1 \cdot 10^6$ & $\mathbf{2.6 \cdot 10^6}$ & $2.2 \cdot 10^6$ & $2.1 \cdot 10^6$ \\
    Australian & $1015$ & $\mathbf{5000}$ & $4234$ & $4058$ & $0.9 \cdot 10^6$ & $\mathbf{2.5 \cdot 10^6}$ & $2.1 \cdot 10^6$ & $2.0 \cdot 10^6$ \\ 
    \midrule
    {Target}&{L2HMC\footnotemark[\value{footnote}]}&{AR}(ours)&{ARLB}(ours)&{VI}&{L2HMC\footnotemark[\value{footnote}]}&{AR}(ours)&{ARLB}(ours)&{VI}\\
    \midrule
    50d ICG & $783$ & $\mathbf{1000}$ & $891$ & $900$ & $2.2\cdot 10^5$ & $\mathbf{5.7\cdot 10^5}$ & $5.1\cdot 10^5$ & $5.1\cdot 10^5$ \\
    RoughWell & $625$ & $\mathbf{1000}$ & $\mathbf{1000}$ & $\mathbf{1000}$ & $1.7\cdot 10^5$ & $\mathbf{5.7\cdot 10^5}$ & $\mathbf{5.7\cdot 10^5}$ & $\mathbf{5.7\cdot 10^5}$ \\
    2d SCG & $497$ & $\mathbf{1000}$ & $\mathbf{1000}$ & $\mathbf{1000}$ & $1.4\cdot 10^5$ & $\mathbf{5.7\cdot 10^5}$ & $\mathbf{5.7\cdot 10^5}$ & $\mathbf{5.7\cdot 10^5}$ \\
    MoG & $32$ & $\mathbf{885}$ & $868$ & $727$ & $0.08 \cdot 10^5$ & $\mathbf{4.8\cdot 10^5}$ & $4.7\cdot 10^5$ & $3.9\cdot 10^5$ \\
    \bottomrule
\end{tabular}}
\end{small}
\end{center}
\vskip -0.1in
\end{table}
\footnotetext{ESS values are taken from the corresponding papers. 
ESS per second is evaluated for the same GPU and framework by reproducing architectures from the papers.}

Here we consider the case when the unnormalized density of a target distribution is given.
For an independent proposal, we take the RealNVP model \citep{dinh2016density} (see details in Appendix \ref{app:nvp}) and learn it via Algorithm \ref{alg:db}, maximizing either the acceptance rate (\textbf{AR}) or its lower bound (\textbf{ARLB}).
To demonstrate the importance of the forward (mass-covering) KL divergence in our objective, we provide a comparison against the variational inference (\textbf{VI}) \citep{de2001variational} for the same architecture of the proposal.
For comparison with HMC-like techniques, we take \textbf{A-NICE-MC} \citep{song2017nice} and \textbf{L2HMC} \citep{levy2017generalizing} algorithms, that outperform the vanilla HMC algorithm.
We evaluate the performance of all samplers by \textit{Effective Sample Size} (ESS) and the ESS per second as suggested in \citep{song2017nice, levy2017generalizing} (see Appendix \ref{app:ess} for formulation).

In Table \ref{tab:results_db}, we show that maximization of the acceptance rate (AR column) outperforms other models in most cases.
As target distributions, we consider different synthetic densities and the posterior of Bayesian logistic regression on different datasets (see Appendix \ref{app:pdfs} for formulation).
Comparing the columns ARLB and VI, we see that adding the forward (mass-covering) KL divergence into the objective is crucial for the sampling.
The most contrast example is Mog6 distribution for which variational inference fails to cover all modes (see Fig. \ref{fig:mog6_hist2d_paper}). 
In Appendix \ref{app:proposals}, we present the examples of learned proposals and histograms of the MH samples.

As suggested in \cite{de2001variational}, we try to mix the proposal learned by the variational inference with a random walk kernel.
However, in our experiments, such mixing does not improve the ESS (and the ESS per second as well).
This fact agrees with the intuition provided in \cite{de2001variational}.
That is, the variational approximation convergences quickly to the regions of high target density and the random walk kernel describes neighborhood of these regions.
Mixing the proposal with a random walk kernel does not improve the ESS since this metric depends on statistics of the whole target distribution.
However, the random walk kernel allows for more accurate statistics estimation of individual components (see Appendix \ref{app:rw}).

Also, we analyze empirically whether the optimization of the acceptance rate lower bound leads to the maximization of the acceptance rate itself.
On a toy example, we show that the loss landscape for the acceptance rate has the same local minima as for its lower bound (Appendix \ref{app:toy}).
During the maximization of the lower bound on synthetic distributions, we observe a high correlation between these objectives (Appendix \ref{app:LB_optimization}).

\begin{SCfigure}[][h]
    \centering
    \includegraphics[width=0.5\textwidth]{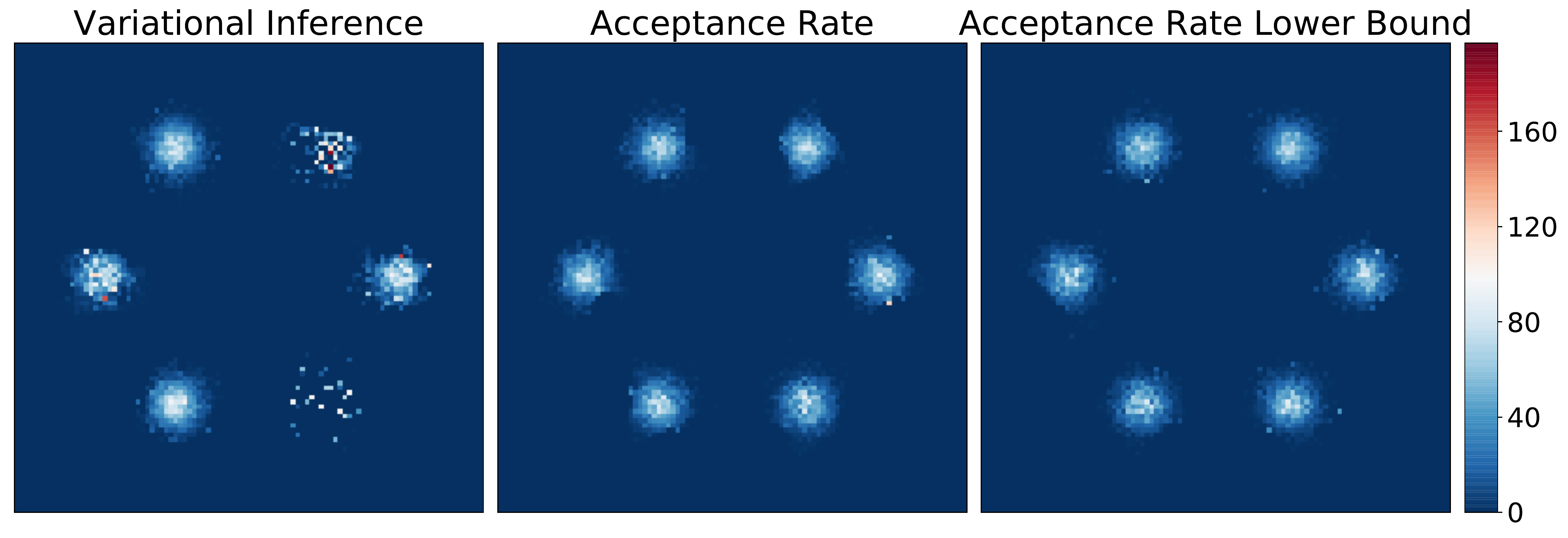}
    \caption{2d histrograms of $25$k samples from the MH algorithm with different proposals.
    From left to right proposals are learned by the variational inference, the acceptance rate maximization, the acceptance rate lower bound maximization.
    }
    \label{fig:mog6_hist2d_paper}
\end{SCfigure}

\subsection{Sample-based setting}
\label{sec:exps_sb}

\begin{table}[t]
\def\arraystretch{1.5}
\caption{Comparison of sampling using the MH algorithm and using the generator for different models. 
Low FID and high IS are better.
For a single evaluation of metrics on CIFAR-10 and CelebA datasets, we use 10k samples, and on ImageNet, we use 50k samples. 
Then we average all the values across 5 independent runs. 
See the description of models in the text.}
\label{tab:gans}
\vskip -0.15in
\begin{center}
\begin{small}
\resizebox{\textwidth}{!}{
\begin{tabular}{lccccccccc}
    \toprule
    \multicolumn{2}{c}{} & \multicolumn{2}{c}{\bf DCGAN} & \multicolumn{2}{c}{\bf WPGAN} & \multicolumn{2}{c}{\bf ARLB} & \multicolumn{1}{c}{\bf BigGAN-100k} & \multicolumn{1}{c}{\bf BigGAN-138k}\\
    \midrule
    & & CIFAR-10 & CelebA & CIFAR-10 & CelebA & CIFAR-10 & CelebA & ImageNet & ImageNet \\
    \midrule
    \multirow{2}{*}{FID} & Generator & $49.06\pm0.34$ & $14.91\pm0.16$ & $47.38\pm0.21$ & $39.09\pm0.33$ & $46.55\pm0.21$& $17.25\pm0.07$ & $11.74\pm 0.06$ & $9.92\pm 0.06$ \\
     & MH(ours) & $\mathbf{46.12}\pm0.29$ & $\mathbf{12.70}\pm0.13$ & $\mathbf{36.65}\pm0.28$ & $\mathbf{25.41}\pm0.28$ & $\mathbf{45.71}\pm0.46$ & $\mathbf{16.57}\pm0.16$ & $\mathbf{10.80}\pm0.04$ & $\mathbf{9.52}\pm0.04$ \\ 
    \midrule 
    \multirow{2}{*}{IS} & Generator & $3.64\pm0.02$ & $2.51\pm0.01$ & $3.52\pm0.02$ & $2.05\pm0.01$ & $3.59\pm0.01$ & $2.38\pm0.02$ & $74.03\pm 0.74$ & $97.73\pm 0.55$ \\
     & MH(ours) & $\mathbf{3.86}\pm0.06$ & $\mathbf{2.73}\pm0.01$ & $\mathbf{4.02}\pm0.03$ & $\mathbf{2.54}\pm0.01$ & $\mathbf{3.72}\pm0.04$ & $\mathbf{2.47}\pm0.01$ & $\mathbf{82.10}\pm0.56$ & $\mathbf{105.62}\pm0.74$ \\
    \bottomrule
\end{tabular}}
\end{small}
\end{center}
\end{table}

Using the Metropolis-Hastings algorithm, we improve the performance of various GANs compared to the straightforward sampling from the generator.
To run the MH algorithm, we treat the generator as a proposal distribution and filter samples from the generator approximating the acceptance test by the discriminator \eqref{eq:approx_test}.

We validate the proposed method on four different models: \textbf{DCGAN} \citep{radford2015unsupervised}, Wasserstein GAN with gradient penalty (\textbf{WPGAN}) \citep{gulrajani2017improved}, recently proposed large scale model \textbf{BigGAN} \citep{brock2018large} and Algorithm \ref{alg:sb} that maximizes the acceptance rate lower bound (\textbf{ARLB}).
For DCGAN, WPGAN, ARLB, we use the same architectures of both generator and discriminator networks.
For BigGAN, we use two checkpoints from the author's repository: one is taken well before collapse (100k generator iterations), and another is taken just before collapse (138k generator iterations).
We also reset the last linear layer of the discriminator in WPGAN and BigGAN and learn it by the minimization of the binary cross-entropy to estimate the density ratio according to \eqref{eq:disc_opt}.

For the performance evaluation, we use the Inception Score (IS) \citep{salimans2016improved} and Frechet Inception Distance (FID) \citep{heusel2017gans}. 
Note that these metrics rely heavily on the implementation of Inception network \citep{barratt2018note}; therefore, for all experiments, we use PyTorch version of the Inception V3 network \citep{paszke2017automatic}.

In Table \ref{tab:gans}, we demonstrate empirical gains for all models using the MH algorithm.
Although we still do not sample from the target distribution, the proposed method alleviates the non-optimality of the generator.
One could expect a perfect match of the target distribution if the discriminator overfits on the target dataset; however, in this case, the acceptance rate will be infeasible.
In our experiments, during sampling via MH, the empirical acceptance rates are approximately $10\%$.
Also, we show that ARLB has comparable performance to the DCGAN, providing the empirical evidence that minimization of the loss \eqref{eq:gan_lb} is equivalent to the conventional GAN training.

\section{Conclusion}
\label{sec:future}

We propose to use the acceptance rate of the MH algorithm as the objective for learning an independent proposal distribution to obtain an efficient sampler.
Compared to the variational inference, this procedure takes the forward KL divergence into account, thus fostering mass-covering.
For empirical target distributions, the proposed method is equivalent to the training of the conventional GAN.
However, application of the MH algorithm allows for more accurate sampling on the test stage.


\bibliography{nips2019}
\bibliographystyle{icml2019}

\newpage

\appendix
\section{Acceptance rate of the MH algorithm}

\subsection{Proof of Theorem 1}
\label{app:proof}

Remind that we have random variables $\xi = \frac{p(x')q(x\cond x')}{p(x)q(x'\cond x)}, x \sim p(x), x' \sim q(x'\cond x)$ and $u \sim \text{Uniform}[0,1]$, and want to prove the following equalities.
\begin{equation}
    \mean_\xi \min\{1, \xi\} = \mathbb{P}\{\xi > u\} = 1 - \frac{1}{2} \mean_\xi |\xi - 1|
\end{equation}
Equality $\mean_\xi \min\{1, \xi\} = \mathbb{P}\{\xi > u\}$ is obvious.
\begin{equation}
    \mean_\xi \min\{1, \xi\} = \int_{0}^\infty p_\xi(x) \min\{1, x\} dx = \int_{x \geq 1} p_\xi(x) dx + \int_{x < 1} p_\xi(x) x dx
\end{equation}
\begin{equation}
    \mathbb{P}\{\xi > u\} = \int_{0}^\infty dx p_\xi(x)\int_0^{x} [0 \leq u \leq 1] du = \int_{x \geq 1}  p_\xi(x)dx + \int_{x < 1}p_\xi(x)x dx
\end{equation}

Equality $\mathbb{P}\{\xi > u\} = 1 - \frac{1}{2} \mean_\xi |\xi - 1|$ can be proofed as follows.
\begin{gather}
    \mathbb{P}\{\xi > u\} = \int_0^1du\int_u^{+\infty} p_{\xi}(x) dx = \int_0^1(1-F_{\xi}(u)) du = \\
    = 1 - \bigg[uF_{\xi}(u)\bigg|_{0}^1 - \int_0^1 up_\xi(u)du \bigg] = 1 - F_{\xi}(1) + \int_0^1up_{\xi}(u)du,
    \label{eq:ar_prob}
\end{gather}
where $F_\xi(u)$ is the CDF of random variable $\xi$. Note that $F_\xi(0) = 0$ since $\xi \in (0,+\infty]$. \eqref{eq:ar_prob} can be rewritten in two ways.
\begin{equation}
    1 - F_{\xi}(1) + \int_0^1up_{\xi}(u)du = 1 + \int_0^1(u-1)p_{\xi}(u)du = 1 - \int_0^1|u-1|p_{\xi}(u)du
    \label{eq:first_way}
\end{equation}
To rewrite \eqref{eq:ar_prob} in the second way we note that $\mean \xi = 1$.
\begin{equation}
    1 - F_{\xi}(1) + \int_0^1up_{\xi}(u)du = \int_{1}^{+\infty}p_\xi(u)du + 1 - \int_1^{+\infty}up_{\xi}(u)du = 1 - \int_{1}^{+\infty}|u-1|p_{\xi}(u)du
    \label{eq:second_way}
\end{equation}
Summing equations \ref{eq:first_way} and \ref{eq:second_way} results in the following formula
\begin{equation}
    \mathbb{P}\{\xi > u\} = 1 - \frac{1}{2} \mean_\xi |\xi - 1|.
\end{equation}
Using the form of $\xi$ we can rewrite the acceptance rate as
\begin{equation}
    1 - \frac{1}{2} \mean_\xi |\xi - 1| = 1 - \mathrm{TV}\bigg(p(x')q(x\cond x') \bigg\Vert p(x)q(x'\cond x)\bigg).
\end{equation}

\subsection{On collapsing to the delta-function}
\label{app:independent}

We first demonstrate that a Markov proposal could achieve the maximal acceptance rate value by collapsing to the delta-function.
We consider the random walk proposal $q(x'\cond x) = \Normal(x'\cond x, \sigma^2)$, then
\begin{equation}
    \mathrm{AR} =  \int dxdx' p(x)\Normal(x'\cond x, \sigma^2) \min\bigg\{1, \frac{p(x')}{p(x)}\bigg\},
\end{equation}
since $\Normal(x'\cond x, \sigma^2) = \Normal(x\cond x', \sigma^2)$.
Now it is clear that we can achieve arbitrary high acceptance rate taking $\sigma$ small enough:
\begin{align}
    \lim_{\sigma \to 0} \mathrm{AR} = \int dxdx' p(x)\delta(x'-x) \min\bigg\{1, \frac{p(x')}{p(x)}\bigg\} = 1.
\end{align}

\begin{figure}[h!]
    \centering
    \includegraphics[width=\textwidth]{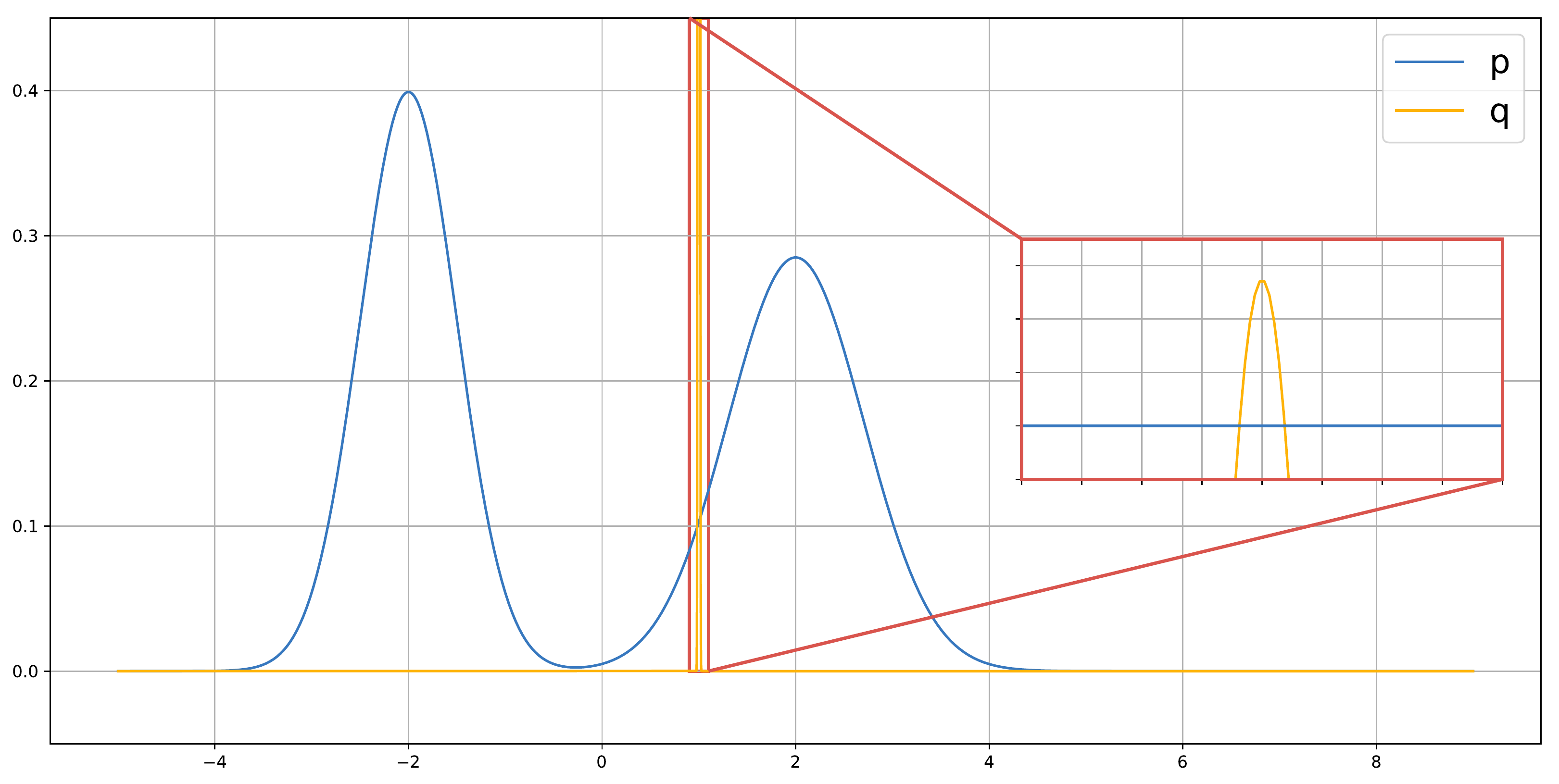}
    \caption{In this figure we show schematic view of approximation of of target distribution with uniform distribution.}
    \label{fig:non_collapsing}
\end{figure}

In the case of the independent proposal, we don't have the collapsing to the delta-function problem.
We provide the intuition for the symmetric KL divergence, but the same holds for the total variation distance.
We consider one-dimensional case where we have some target distribution $p(x)$ and independent proposal $q(x) = \Normal (x \cond \mu, \sigma)$.
Choosing $\sigma$ small enough, we approximate the sampling by MH as sampling on some finite support $x \in  [\mu - a, \mu + a]$.
For this support, we approximate the target distribution with the uniform distribution (see Fig. \ref{fig:non_collapsing}).

For such approximation, optimization of lower bound takes the form
\begin{gather}
    \min_q\bigg[\text{KL}(p(x)\Vert q(x)) + \text{KL}(q(x)\Vert p(x))\bigg]\\
    \min_\sigma \bigg[\text{KL}(\text{Uniform}[-a,a]\Vert \Normal (x \cond 0, \sigma, -a, a)) + \text{KL}(\Normal (x \cond 0, \sigma, -a, a)\Vert\text{Uniform}[-a,a])\bigg]
\end{gather}
Here $\Normal (x \cond 0, \sigma, -a, a)$ is truncated normal distribution. 
The first KL-divergence can be written as follows.
\begin{gather}
    \text{KL}(\text{Uniform}[-a,a]\Vert \Normal (x \cond 0, \sigma, -a, a)) = -\frac{1}{2a}\int_{-a}^a dx \log \Normal (x \cond 0, \sigma, -a, a) - \log 2a = \\
    = -\frac{1}{2a} \bigg[-2a\log (\sigma Z) - a \log 2\pi -\frac{1}{2\sigma^2} \frac{2a^3}{3} \bigg] - \log 2a =\\
    = \log \sigma  + \log Z +\frac{a^2}{6\sigma^2} + \frac{1}{2}\log 2\pi - \log 2a
\end{gather}
Here $Z$ is normalization constant of truncated log normal distribution and $Z=\Phi(a/\sigma) - \Phi(-a/\sigma)$, where $\Phi(x)$ is CDF of standard normal distribution. 
The second KL-divergence is
\begin{gather}
    \text{KL}(\Normal (x \cond 0, \sigma, -a, a)\Vert \text{Uniform}[-a,a]) = \\
    = -\frac{1}{2}\log(2\pi e) - \log\sigma - \log Z + \frac{a}{ \sqrt{2\pi} \sigma Z}\exp\bigg(-\frac{a^2}{2\sigma^2}\bigg) +  \log 2a
\end{gather}
Summing up two KL-divergencies and taking derivative w.r.t. $\sigma$ we obtain
\begin{gather}
    \frac{\partial}{\partial \sigma}\bigg( \text{KL}(\text{Uniform}[-a,a]\Vert \Normal (x \cond 0, \sigma, -a, a)) + \text{KL}(\Normal (x \cond 0, \sigma, -a, a)\Vert \text{Uniform}[-a,a]) \bigg) = \\
    = -\frac{a^2}{3\sigma^3} + \frac{a^3}{\sqrt{2\pi}\sigma^4 Z} \exp\bigg(-\frac{a^2}{2\sigma^2}\bigg) + \frac{a}{\sqrt{2\pi}} \exp\bigg(-\frac{a^2}{2\sigma^2}\bigg)\bigg[ -\frac{1}{\sigma^2 Z} - \frac{1}{\sigma Z^2} \frac{-2a}{\sigma^2\sqrt{2\pi}} \exp\bigg(-\frac{a^2}{2\sigma^2}\bigg) \bigg] = \\
    = \frac{1}{a}\bigg[-\frac{a^3}{3\sigma^3} + \frac{a^2}{\sqrt{2\pi}\sigma^2Z} \exp\bigg(-\frac{a^2}{2\sigma^2}\bigg) \bigg( \frac{a^2}{\sigma^2} -1 + \frac{2a}{\sqrt{2\pi}\sigma Z} \exp\bigg(-\frac{a^2}{2\sigma^2}\bigg) \bigg) \bigg]
\end{gather}
To show that the derivative of the lower bound w.r.t. $\sigma$ is negative, we need to prove that the following inequality holds for positive $x$.
\begin{equation}
    -\frac{1}{3}x^3 + \frac{x^2}{\sqrt{2\pi}(\Phi(x)-\Phi(-x))} \exp(-x^2/2) \bigg( x^2 -1 + \frac{2x}{\sqrt{2\pi} (\Phi(x)-\Phi(-x))} \exp(-x^2/2) \bigg) < 0, \;\; x > 0
    \label{eq:xy}
\end{equation}
Defining $\phi(x) = \int_0^x e^{-t^2/2}dt$ and noting that $2\phi(x) = \sqrt{2\pi} (\Phi(x)-\Phi(-x)) $ we can rewrite inequality \ref{eq:xy} as
\begin{equation}
    \frac{1}{\phi(x)} e^{-x^2/2} \bigg( x^2 -1 + \frac{2xe^{-x^2/2}}{\phi(x)}  \bigg) < \frac{2x}{3}, \;\; x > 0
\end{equation}

By the fundamental theorem of calculus, we have
\begin{equation}
    xe^{-x^2/2} = \int_{0}^x e^{-t^2/2}(1-t^2)dt
\end{equation}
Hence,
\begin{equation}
    \phi(x) - xe^{-x^2/2} = \int_{0}^x e^{-t^2/2}t^2dt \geq  e^{-x^2/2} \int_{0}^x t^2dt = e^{-x^2/2} \frac{x^3}{3}
\end{equation}
Or equivalently, 
\begin{equation}
    \phi(x) \geq  e^{-x^2/2} \frac{x^3+3x}{3}
\end{equation}
Using this inequality twice, we obtain
\begin{equation}
    \frac{e^{-x^2/2}}{\phi(x)} \leq  \frac{3}{x(x^2+3)}
\end{equation}
and
\begin{equation}
    x^2 - 1 + \frac{xe^{-x^2/2}}{\phi(x)} \leq  x^2 - 1 + \frac{3}{x^2+3} = \frac{x^2(2+x^2)}{x^2+3}
\end{equation}

Thus, the target inequality can be verified by the verification of 
\begin{equation}
    \frac{3x(2+x^2)}{(x^2+3)^2} \leq \frac{2x}{3}.
\end{equation}



Thus, we show that partial derivative of our lower bound w.r.t. $\sigma$ is negative.
Using that knowledge we can improve our loss by taking a bigger value of $\sigma$.
Hence, such proposal does not collapse to delta-function.

\subsection{The acceptance rate of independent MH defines a semimetric in distribution space}
\label{app:semi}
In the independent case, we have $\xi = \frac{p(x')q(x)}{p(x)q(x')}, x \sim p(x), x' \sim q(x')$ and we want to prove that $\mean_\xi |\xi - 1|$ is semimetric (or pseudo-metric) in the space of distributions.
In this appendix, we denote $D(p,q) = \mean_\xi |\xi - 1|$.
The first two axioms for metric obviously hold
\begin{enumerate}
    \item $D(p,q) = 0 \iff p = q$
    \item $D(p,q) = D(q,p)$
\end{enumerate}
There is an example when triangle inequality does not hold.
For distributions $p = \text{Uniform}[0, 2/3], q = \text{Uniform}[1/3, 1], s = \text{Uniform}[0, 1]$
\begin{equation}
    D(p,s) + D(q,s) = \frac{4}{3} < \frac{3}{2} = D(p,q).
\end{equation}

But weaker inequality can be proved.
\begin{gather}
    D(p,s) + D(q,s) = \int |p(x)s(y)-p(y)s(x)|dydx + \int |q(x)s(y)-q(y)s(x)|dydx = \\
    = \int \bigg[|\underbrace{p(x)s(y)q(z)}_a - \underbrace{p(y)s(x)q(z)}_b| + |\underbrace{q(x)s(y)p(z)}_c-\underbrace{q(y)s(x)p(z)}_d|  \bigg]dxdydz
    \label{eq:sm_first}
\end{gather}
\begin{gather}
    D(p,s) + D(q,s) = \int |p(z)s(y)q(x)-p(y)s(z)q(x)|dxdydz + \\
     + \int |q(x)s(z)p(y)-q(z)s(x)p(y)|dxdydz \geq
    \int \bigg|\underbrace{q(x)s(y)p(z)}_c - \underbrace{p(y)s(x)q(z)}_b\bigg|dxdydz
    \label{eq:sm_second}
\end{gather}
\begin{gather}
    D(p,s) + D(q,s) = \int |p(z)s(x)q(y)-p(x)s(z)q(y)|dxdydz + \\
     + \int |q(y)s(z)p(x)-q(z)s(y)p(x)|dxdydz \geq
    \int \bigg|\underbrace{q(y)s(x)p(z)}_d - \underbrace{p(x)s(y)q(z)}_a\bigg|dxdydz
    \label{eq:sm_third}
\end{gather}

Summing up equations \ref{eq:sm_first}, \ref{eq:sm_second} and \ref{eq:sm_third}, we obtain
\begin{gather}
    3(D(p,s) + D(q,s)) \geq \int dxdydz \bigg[ |a-b| + |c-d| + |c-b| + |d - a|\bigg] \geq 2\int dxdydz |d-b| = \\
    = 2\int dxdydz s(x)\bigg|q(y)p(z) - q(z)p(y)\bigg| = 2D(p,q)\\
    D(p,s) + D(q,s) \geq \frac{2}{3}D(p,q)
\end{gather}

\section{Optimization of proposal distribution}

\subsection{Learning a discriminator for Markov proposal}
\label{app:markov_gan}

In this section we show how to learn a discriminator in the case of a Markov proposal.
The loss function $\mathcal{L}(\phi)$ for Markov proposal takes form
\begin{equation}
    \mathcal{L}(\phi) = -\text{\Large $\mean$}_{\text{\footnotesize $\begin{matrix}
    x \sim p(x) \\ x' \sim q_\phi(x'\cond x)\end{matrix}$}} l\bigg(\frac{p(x')q_\phi(x\cond x')}{p(x)q_\phi(x'\cond x)}\bigg).
\end{equation}
To estimate ratio $\frac{p(x)q_\phi(x'\cond x)}{p(x')q_\phi(x\cond x')}$  we suggest to use well-known technique of density ratio estimation via training discriminator network.
Denoting discriminator output as $D(x,x')$, we suggest the following optimization problem for the discriminator.
\begin{equation}
    \min_D \bigg[-\text{\Large $\mean$}_{\text{\footnotesize $\begin{matrix}
    x \sim p(x) \\ x' \sim q_\phi(x'\cond x)\end{matrix}$}} \log D(x,x') -\text{\Large $\mean$}_{\text{\footnotesize $\begin{matrix}
    x \sim p(x) \\ x' \sim q_\phi(x'\cond x)\end{matrix}$}} \log (1-D(x',x)) \bigg]
    \label{eq:disc_opt_markov}
\end{equation}
Speaking informally, such discriminator takes two images as input and tries to figure out which image is sampled from true distribution and which one is generated by the one step of proposal distribution.
It is easy to show that the optimal discriminator in the problem \eqref{eq:disc_opt_markov} will be
\begin{equation}
    D(x,x') = \frac{p(x)q_\phi(x'\cond x)}{p(x)q_\phi(x'\cond x) + p(x')q_\phi(x\cond x')}.
\end{equation}
Note that for optimal discriminator we have $D(x,x') = 1 - D(x',x)$.
In practice, we have no optimal discriminator and these values can differ significantly.
Thus, we have four ways for density ratio estimation that may differ significantly.
\begin{equation}
    \frac{p(x)q_\phi(x'\cond x)}{p(x')q_\phi(x\cond x')} \approx  \frac{D(x,x')}{1-D(x,x')} \approx \frac{1-D(x',x)}{D(x',x)} \approx \frac{1-D(x',x)}{1-D(x,x')} \approx \frac{D(x,x')}{D(x',x)}
\end{equation}
To avoid this ambiguity we suggest to use the discriminator of a special structure.
Let $\widetilde{D}(x,x')$ be a convolutional neural network with scalar output.
Then the output of discriminator $D(x,x')$ is defined as follows.
\begin{equation}
    D(x,x') = \frac{\exp(\widetilde{D}(x,x'))}{\exp(\widetilde{D}(x,x')) + \exp(\widetilde{D}(x',x))}
\end{equation}
In other words, such discriminator can be described as the following procedure.
For single neural network $\widetilde{D}(\cdot, \cdot)$ we evaluate two outputs $\widetilde{D}(x, x')$ and $\widetilde{D}(x', x)$.
Then we take softmax operation for these values.

\subsection{Intuition for better gradients}
\label{app:sb_grads}
In this section, we provide an intuition for sample-based setting that the loss function for lower bound has better gradients than the loss function for acceptance rate.
Firstly, we remind that in the sample-based setting we use a discriminator for density ratio estimation.
\begin{equation}
    D(x,x') = \frac{p(x)q(x'\cond x)}{p(x)q(x'\cond x) + p(x')q(x\cond x')}
\end{equation}
For this purpose we use the discriminator of special structure
\begin{equation}
    D(x,x') = \frac{\exp(\widetilde{D}(x,x'))}{\exp(\widetilde{D}(x,x')) + \exp(\widetilde{D}(x',x))} = \frac{1}{1 + \exp\bigg(-(\widetilde{D}(x,x') - \widetilde{D}(x',x))\bigg)}
\end{equation}
We denote $d(x,x') = \widetilde{D}(x,x') - \widetilde{D}(x',x)$ and consider the case when the discriminator can easily distinguish fake pairs from valid pairs. 
So $D(x,x')$ is close to $1$ and $d(x,x') \gg 0$ for $x \sim p(x)$ and $x' \sim q(x'\cond x)$.
To evaluate gradients we consider Monte Carlo estimations of each loss and take gradients w.r.t. $x'$ in order to obtain gradients for parameters of proposal distribution.
We do not introduce the reparameterization trick to simplify the notation but assume it to be performed.
For the optimization of the acceptance rate we have
\begin{equation}
    \int dxdx' p(x) q(x'\cond x) \bigg| \frac{p(x') q(x\cond x')}{p(x) q(x'\cond x)} - 1\bigg| \simeq \bigg|\frac{p(x') q(x\cond x')}{p(x) q(x'\cond x)} - 1\bigg|
\end{equation}
\begin{gather}
    L_{\text{AR}} = \bigg|\frac{p(x') q(x\cond x')}{p(x) q(x'\cond x)} - 1\bigg| \approx \bigg|\frac{1-D(x,x')}{D(x,x')} - 1\bigg| \\
    \frac{\partial L_{\text{AR}}}{\partial x'} = \frac{1}{D^2(x,x')}\frac{\partial D(x,x')}{\partial x'} = \exp(-d(x,x')) \frac{\partial d(x,x')}{\partial x'}
    \label{eq:ar_grad}
\end{gather}
While for the optimization of the lower bound we have
\begin{equation}
    \int dxdx' p(x) q(x'\cond x) \log\bigg( \frac{p(x) q(x'\cond x)}{p(x') q(x\cond x')} \bigg) \simeq \log\bigg( \frac{p(x) q(x'\cond x)}{p(x') q(x\cond x')} \bigg)
\end{equation}
\begin{gather}
    L_{\text{LB}} = -\log \bigg(\frac{p(x') q(x\cond x')}{p(x) q(x'\cond x)}\bigg) \approx -\log \bigg(\frac{1-D(x,x')}{D(x,x')}\bigg) \\
    \frac{\partial L_{\text{LB}}}{\partial x'} = \frac{1}{(1-D(x,x'))D(x,x')}\frac{\partial D(x,x')}{\partial x'} = \frac{\partial d(x,x')}{\partial x'}
    \label{eq:lb_grad}
\end{gather}
Now we compare \eqref{eq:ar_grad} and \eqref{eq:lb_grad}. 
We see that in case of overconfident discriminator we have vanishing gradients in \eqref{eq:ar_grad} due to $\exp(-d(x,x'))$, while it is not the case for \eqref{eq:lb_grad}. See also Fig. \ref{fig:toy_example} for loss landscapes of these two losses.

\section{Experiments}
\subsection{Effective Sample Size formulation}
\label{app:ess}

For the effective sample size formulation we follow \cite{song2017nice}.

Assume a target distribution $p(x)$, and a Markov chain Monte Carlo (MCMC) sampler that produces a set of N correlated samples $\{x_i\}_{1}^{N}$ from some distribution $q(\{x_i\}_{1}^{N})$ such that $q(x_i) = p(x_i)$. Suppose we are estimating the mean of $p(x)$ through sampling; we assume that increasing the number of samples will reduce the variance of that estimate.

Let $V = \Var_q[\sum_{i=1}^{N} x_i/N]$ be the variance of the mean estimate through the MCMC samples. The effective sample size (ESS) of $\{x_i\}_{1}^{N}$, which we denote as $M = ESS(\{x_i\}_{1}^{N})$, is the number of independent samples from $p(x)$ needed in order to achieve the same variance, i.e. $\Var_p[\sum_{j=1}^{M} x_j / M] = V$. A practical algorithm to compute the ESS given $\{x_i\}_{1}^{N}$ is provided by:

\begin{equation}
ESS(\{x_i\}_{1}^{N}) = \frac{N}{1 + 2 \sum_{s=1}^{N-1}(1 - \frac{s}{N})\rho_s}
\end{equation}
where $\rho_s$ denotes the autocorrelation under $q$ of $x$ at lag $s$. We compute the following empirical estimate $\hat{\rho}_s$ for $\rho_s$:
\begin{equation}
\hat{\rho}_s = \frac{1}{\hat{\sigma}^2 (N - s)} \sum_{n = s+1}^{N} (x_n - \hat{\mu}) (x_{n-s} - \hat{\mu})
\end{equation}
where $\hat{\mu}$ and $\hat{\sigma}$ are the empirical mean and variance obtained by an independent sampler. 

Due to the noise in large lags $s$, we adopt the approach of \cite{hoffman2014no} where we truncate the sum over the autocorrelations when the autocorrelation goes below 0.05.

\subsection{Architecture of the RealNVP proposal}
\label{app:nvp}

For the proposal distribution, we use a similar architecture to the NICE proposal.
The RealNVP model \citep{dinh2016density} use the same strategy for evaluating the Jacobian as the NICE model does.
Each \textit{coupling layer} defines the following function. 
Given a $D$ dimensional input $x$ and $d < D$, the output $y$ is evaluated by the formula
\begin{align*}
    & y_{1:d} = x_{1:d}, \\
    & y_{d+1:D} = x_{d+1:D} \odot \exp(s(x_{1:d}))+ t(x_{1:d}),
\end{align*}
where the functions $s,t$ can be arbitrary complex, since the structure of the functions doesn't influence the computation of the Jacobian.

For our proposal we use $4$ coupling layers with $s$ and $t$ consist of two fully-connected layers with hidden dimension of $512$.

\subsection{Target distribution in the density-based setting}
\label{app:pdfs}

For synthetic distributions we consider the same distributions as in \cite{song2017nice} and \cite{levy2017generalizing}.

The analytic form of $p(x)$ for \textit{ring} is:
\begin{equation}
p(x) \propto \exp(-U(x)), \;\;\; U(x) = \frac{(\sqrt{x_1^2 + x_2^2} - 2)^2}{0.32}
\end{equation}

The analytic form of $p(x)$ for \textit{mog2} is:
\begin{equation}
p(x) = \frac{1}{2}\Normal(x | \mu_1, \sigma_1) + \frac{1}{2}\Normal(x | \mu_2, \sigma_2)
\end{equation}
where $\mu_1 = [5, 0]$, $\mu_2 = [-5, 0]$, $\sigma_1 = \sigma_2 = [0.5, 0.5]$. 

The analytic form of $p(x)$ for \textit{mog6} is:
\begin{equation}
p(x) = \frac{1}{6}\sum_{i=1}^{6} \Normal(x | \mu_i, \sigma_i)
\end{equation}
where $\mu_i = [\sin \frac{i\pi}{3}, \cos \frac{i\pi}{3}]$ and $\sigma_i = [0.5, 0.5]$.

The analytic form of $p(x)$ for \textit{ring5} is:
\begin{equation}
p(x) \propto \exp(-U(x)), \;\;\; U(x) = \min (u_1, u_2, u_3, u_4, u_5)
\end{equation}
where $u_i = (\sqrt{x_1^2 + x_2^2} - i)^2 / 0.04$.

The analytic form of $p(x)$ for \textit{ICG} is:
\begin{equation}
    p(x) = \Normal(0,\Sigma),
\end{equation}
where $\Sigma$ is diagonal matrix with  diagonal spaced  log-linearly between $10^{-2}$ and $10^2$.

The analytic form of $p(x)$ for \textit{SCG} is:
\begin{equation}
    p(x) = \Normal(0,B\Sigma B^T), \;\;\; \Sigma = 
    \begin{bmatrix}
    10^{-2} & 0  \\
    0 & 10^2
    \end{bmatrix}, \;\;\;
    B =
    \begin{bmatrix}
    1/\sqrt{2} & -1/\sqrt{2}  \\
    1/\sqrt{2} & 1/\sqrt{2}
    \end{bmatrix}.
\end{equation}

The analytic form of $p(x)$ for \textit{RoughWell} is:
\begin{equation}
p(x) \propto \exp(-U(x)), \;\;\; U(x) = \frac{1}{2}x^Tx + \eta\sum_i \cos(\frac{x_i}{\eta}), \;\;\; x \in \mathbb{R}^2, \;\;\; \eta = 10^{-2}.
\end{equation}

The analytic form of $p(x)$ for \textit{MoG} is:
\begin{equation}
p(x) = \frac{1}{2}\Normal(x | \mu_1, \sigma_1) + \frac{1}{2}\Normal(x | \mu_2, \sigma_2)
\end{equation}
where $\mu_1 = [2, 0]$, $\mu_2 = [-2, 0]$, $\sigma_1^2 = \sigma_2^2 = [0.1, 0.1]$. 

For the Bayesian logistic regression, we define likelihood and prior as
\begin{equation}
    p(y=1\cond x, \theta) = \frac{1}{1+\exp(-x^T\theta_w+\theta_b)}, \;\;\; p(\theta) = \Normal(\theta\cond 0,1).
\end{equation}
Then the unnormalized density of the posterior distribution for a dataset $D=\{(x_i,y_i)\}_i$ is
\begin{equation}
    p(\theta\cond D) \propto \prod_i p(y_i\cond x_i, \theta) p(\theta).
\end{equation}
We sample from the posterior distribution on three datasets: german ($25$ covariates, $1000$ data points), heart ($14$ covariates, $532$ data points) and australian ($15$ covariates, $690$ data points).

\subsection{Toy problem}
\label{app:toy}

\begin{figure}[H]
    \centering
    \includegraphics[width=0.49\textwidth]{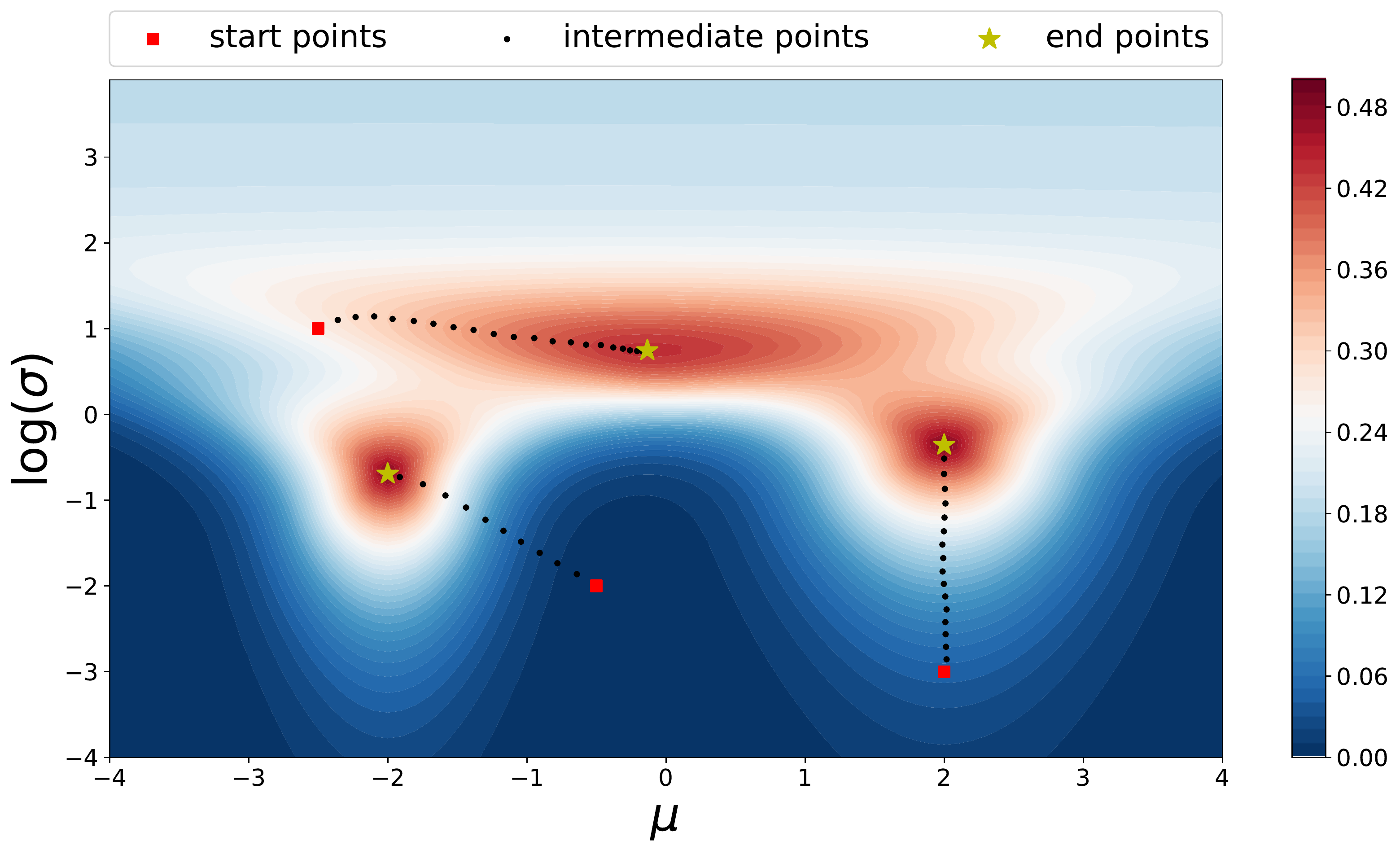}
    \includegraphics[width=0.49\textwidth]{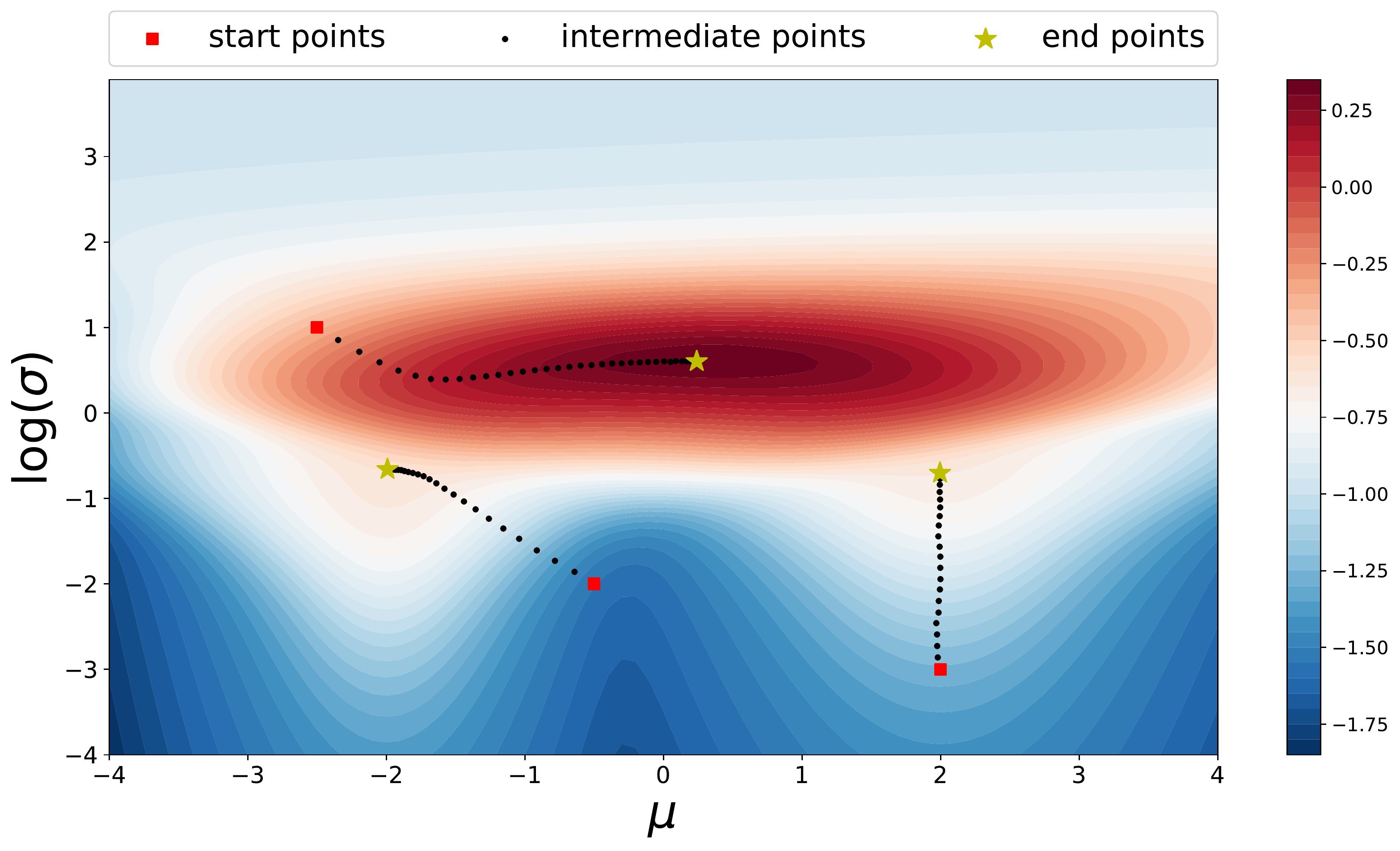}
    \caption{Landscapes in the parameter space for the toy problem.
    Left: level-plot for the acceptance rate of the MH algorithm.
    Right: level-plot for the lower bound of the acceptance rate.
    }
    \label{fig:toy_example}
\end{figure}

This experiment shows that the acceptance rate has a similar landscape to its lower bound.
For the target distribution we consider bimodal Gaussian $p(x) = 0.5\cdot\Normal(x\cond -2, 0.5) + 0.5\cdot\Normal(x\cond 2, 0.7)$, for the independent proposal we consider the gaussian $q(x) = \Normal(x\cond \mu, \sigma)$.
We perform stochastic gradient optimization using Algorithm \ref{alg:db} from the same initialization for both objectives (Fig. \ref{fig:toy_example}) and obtain approximately the same local maximum.

\subsection{Optimization of the lower bound}
\label{app:LB_optimization}
\begin{figure}[h]
\centering
\includegraphics[width=0.49\textwidth]{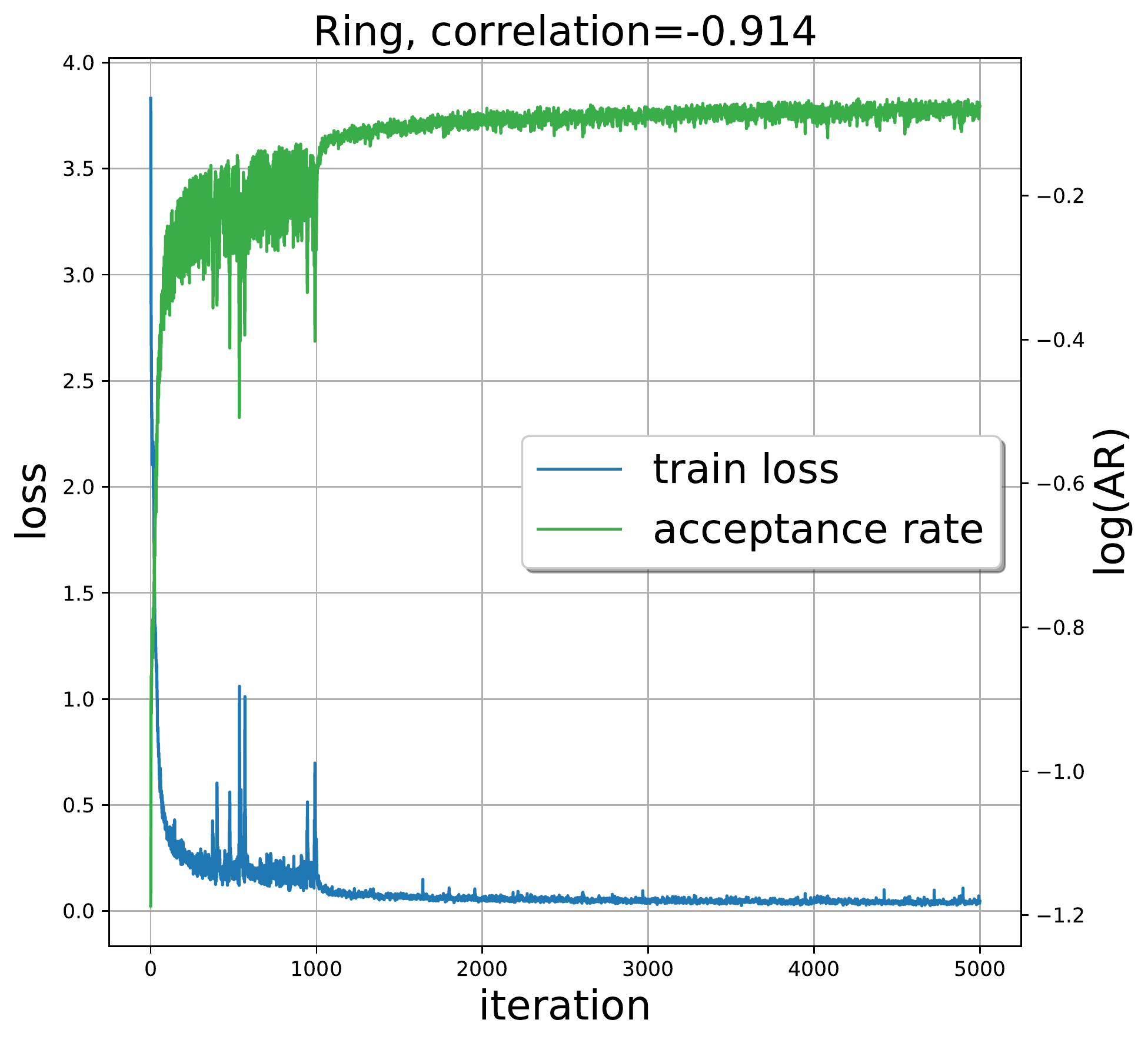}
\includegraphics[width=0.49\textwidth]{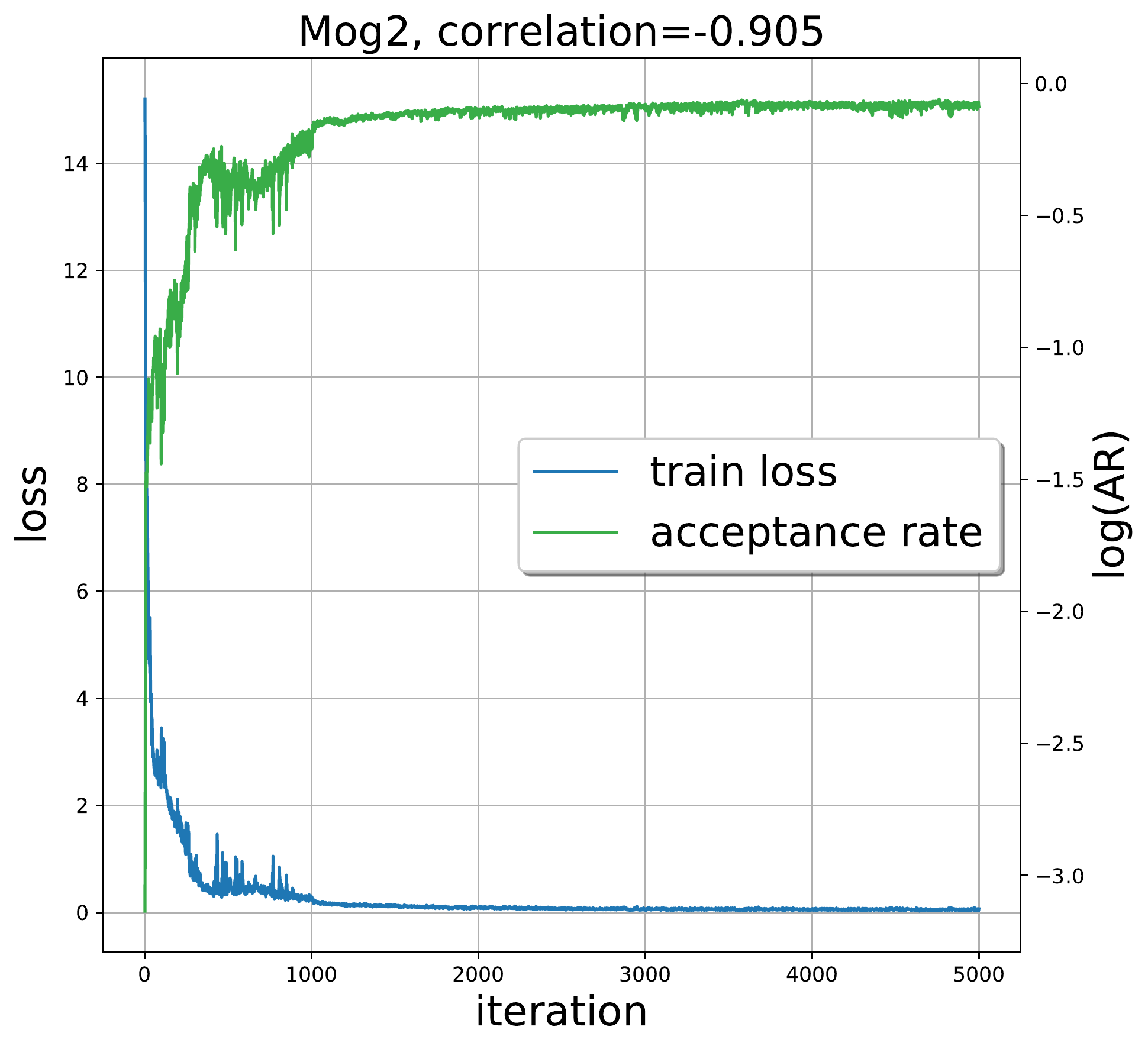}
\\
\includegraphics[width=0.49\textwidth]{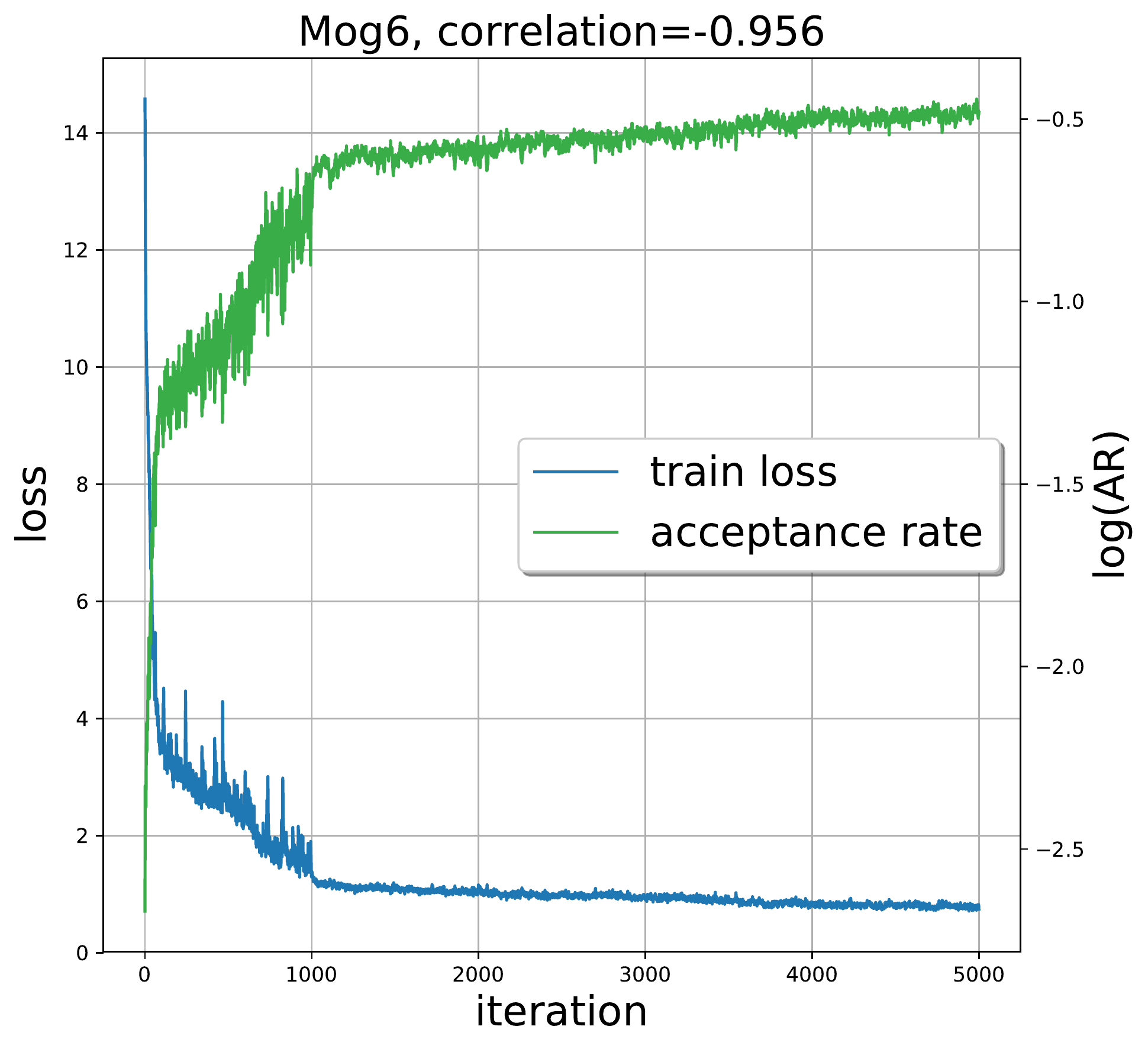}
\includegraphics[width=0.49\textwidth]{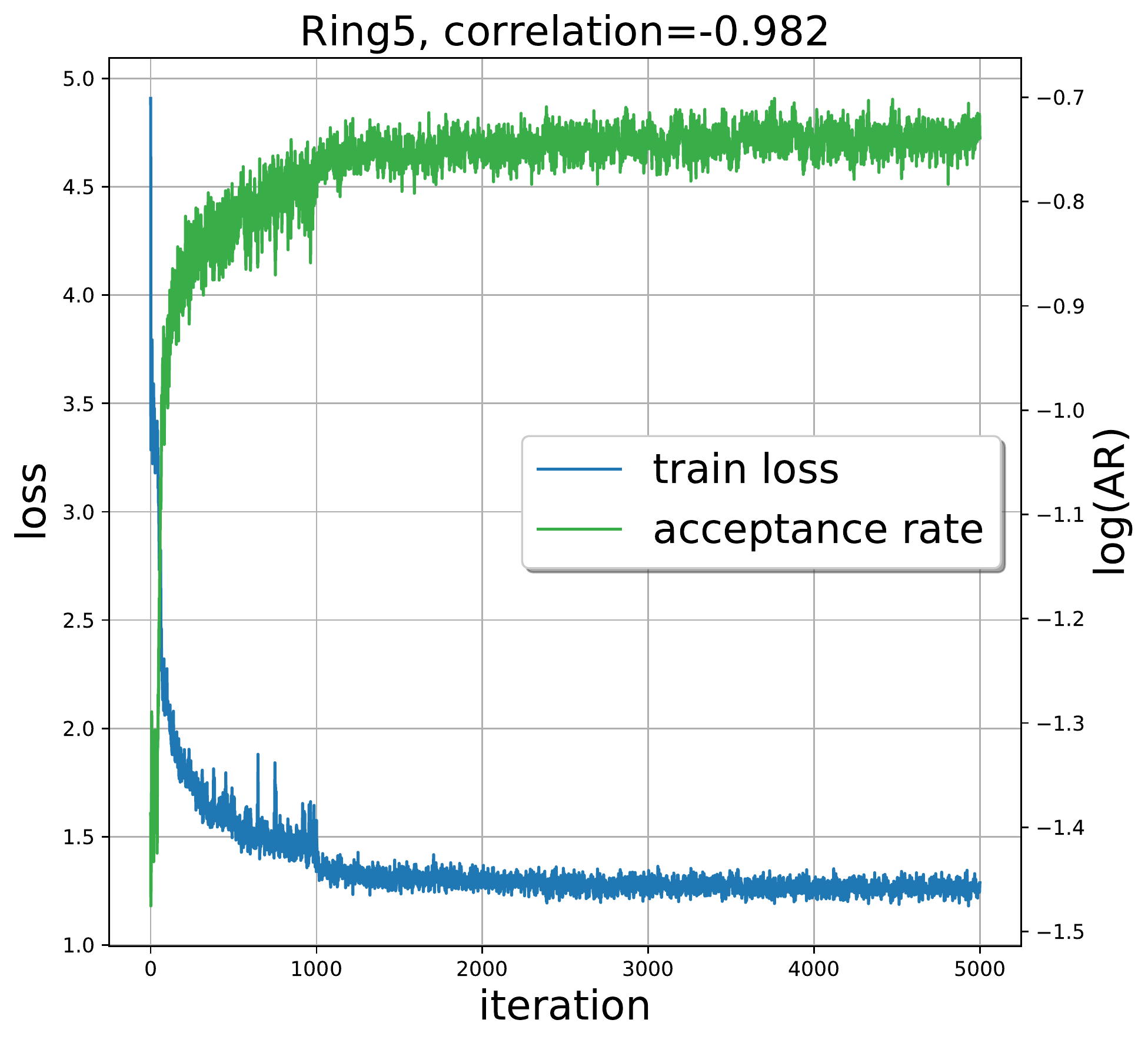}
\caption{
plots for the acceptance rate and the acceptance rate lower bound evaluated at every iteration during the optimization of the acceptance rate lower bound.
Correlation coefficient is evaluated between the logarithm of the acceptance rate and the acceptance rate lower bound.
}
\label{fig:LB_corr}
\end{figure}

In this section we provide the empirical evidence that maximization of the proposed lower bound on the acceptance rate (\textbf{ARLB}) results in maximization of the acceptance rate (\textbf{AR}).
For that purpose we evaluate \textbf{ARLB} and \textbf{AR} at each iteration during the optimization of \textbf{ARLB}.
After training we evaluate correlation coefficient between \textbf{ARLB} and logarithm of \textbf{AR}.
The curves are shown in Fig. \ref{fig:LB_corr}. 
Correlation coefficients for different distributions are: $-0.914$ (\textbf{ring}), $-0.905$ (\textbf{mog2}), $-0.956$ (\textbf{mog6}), $-0.982$ (\textbf{ring5}).

\clearpage

\subsection{Learned proposals}
\label{app:proposals}
\begin{figure}[h]
    \centering
    \includegraphics[width=0.49\textwidth]{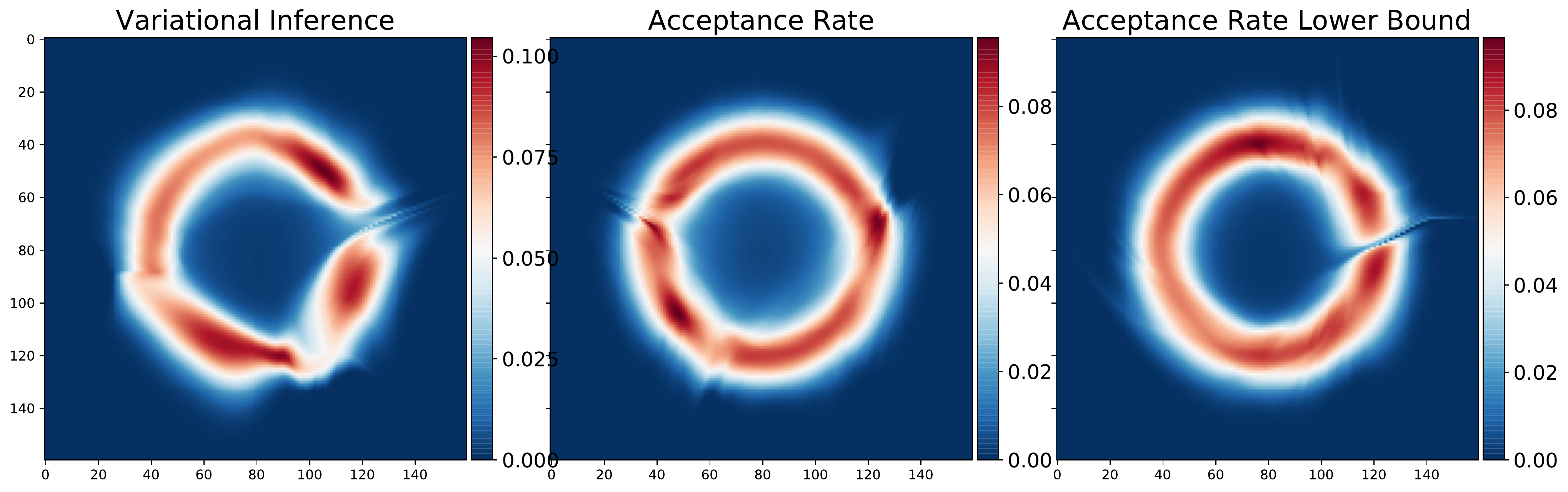}
    \includegraphics[width=0.49\textwidth]{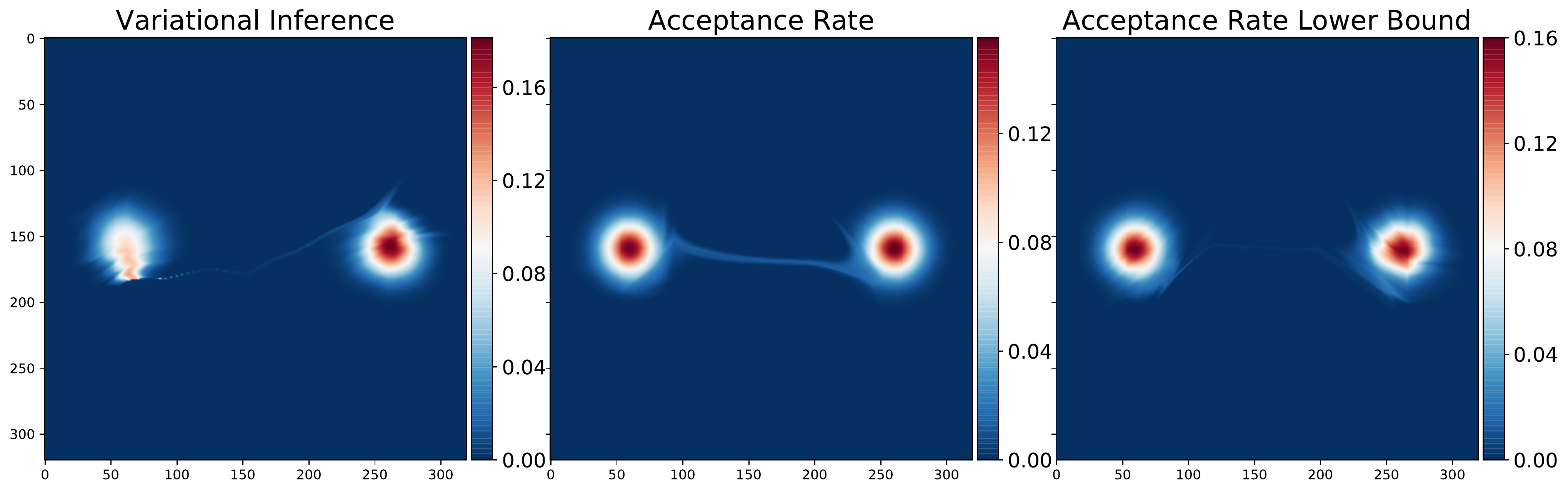}
    \\
    \includegraphics[width=0.49\textwidth]{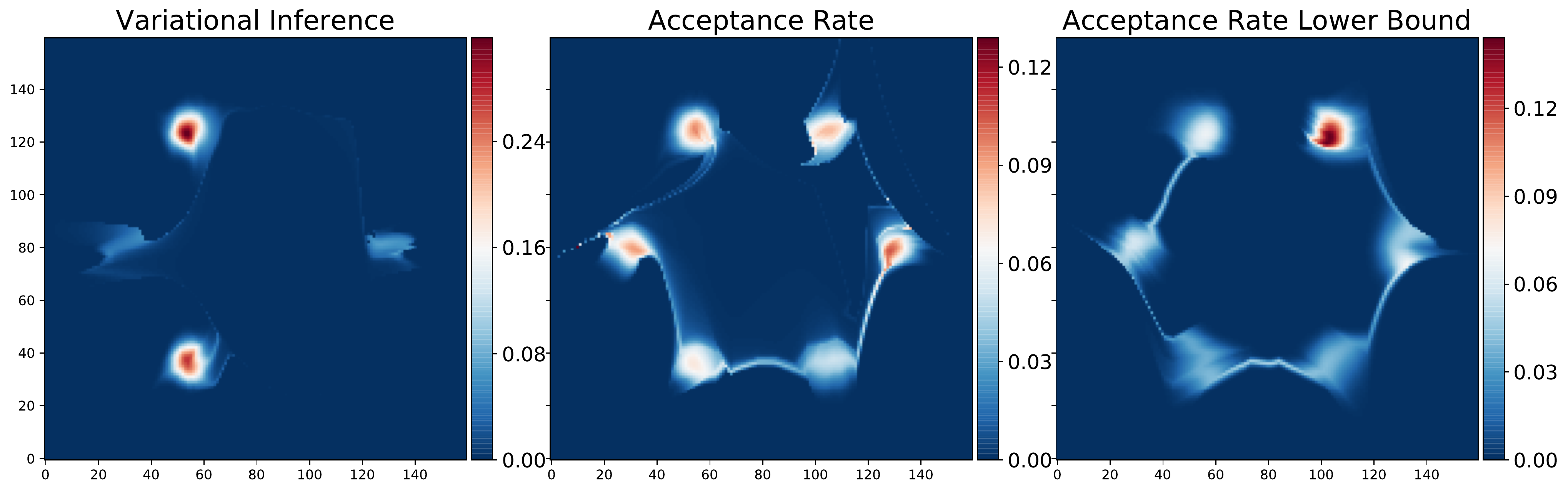}
    \includegraphics[width=0.49\textwidth]{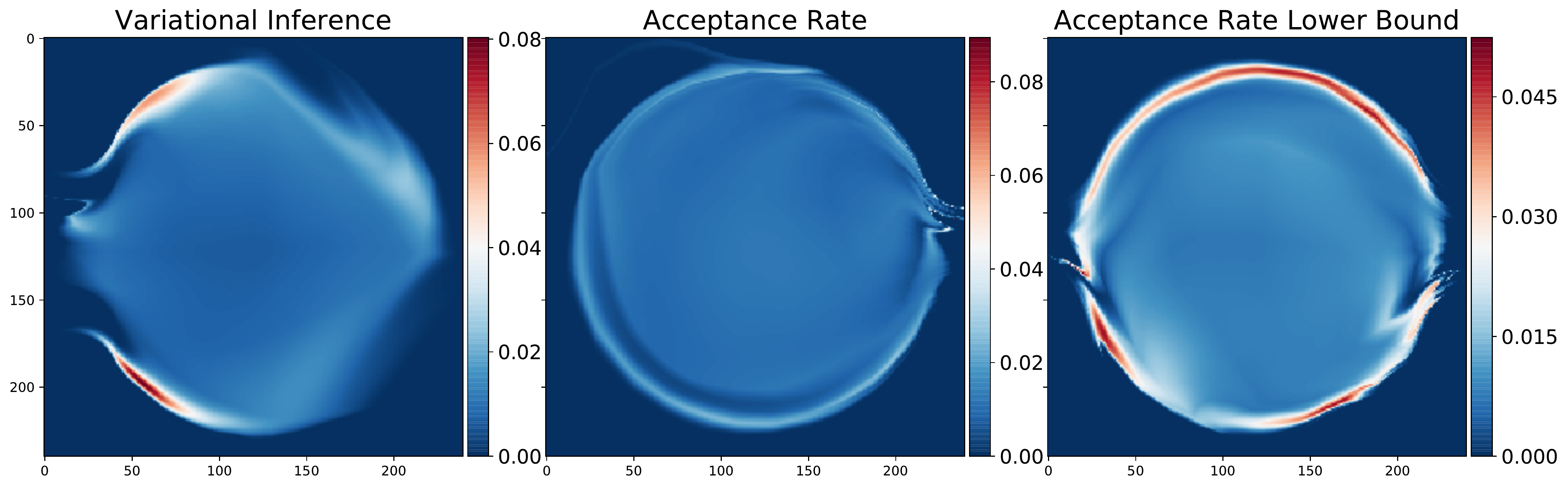}
    \caption{levelplots of learned proposal densities.
    For each distribution from left to right proposals are learned by:  variational inference, the acceptance rate maximization, the acceptance rate lower bound maximization.}
    \label{fig:density}
\end{figure}

\begin{figure}[h]
    \centering
    \includegraphics[width=0.49\textwidth]{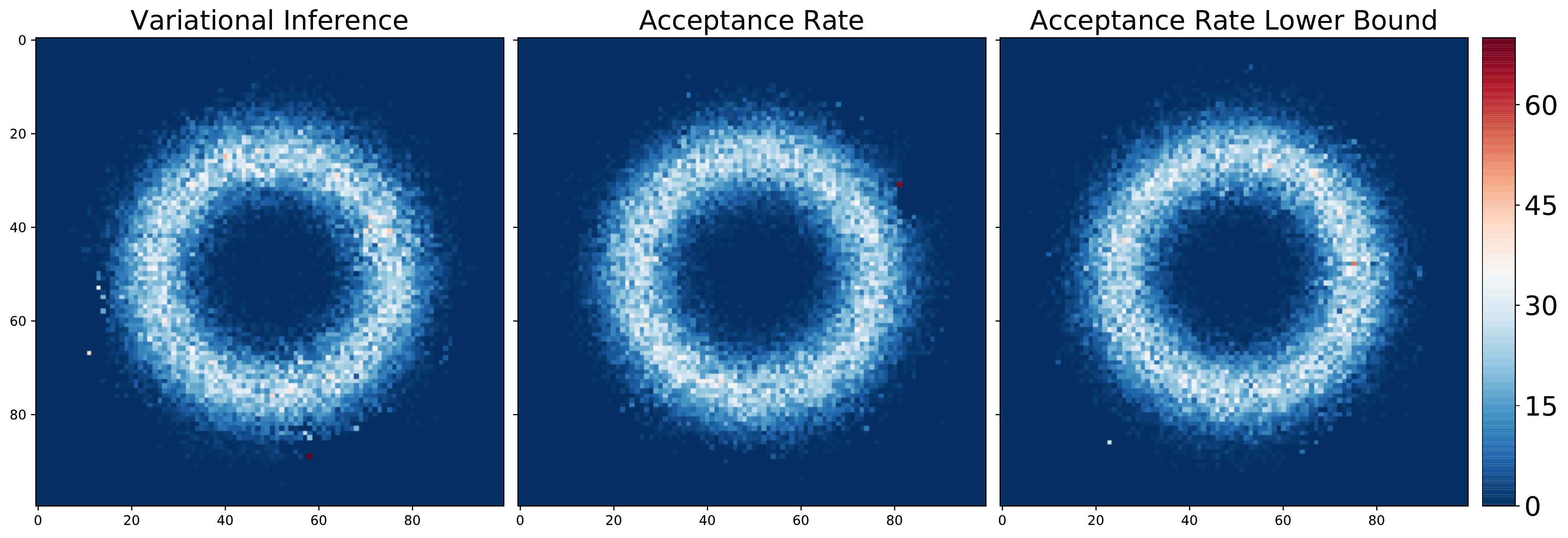}
    \includegraphics[width=0.49\textwidth]{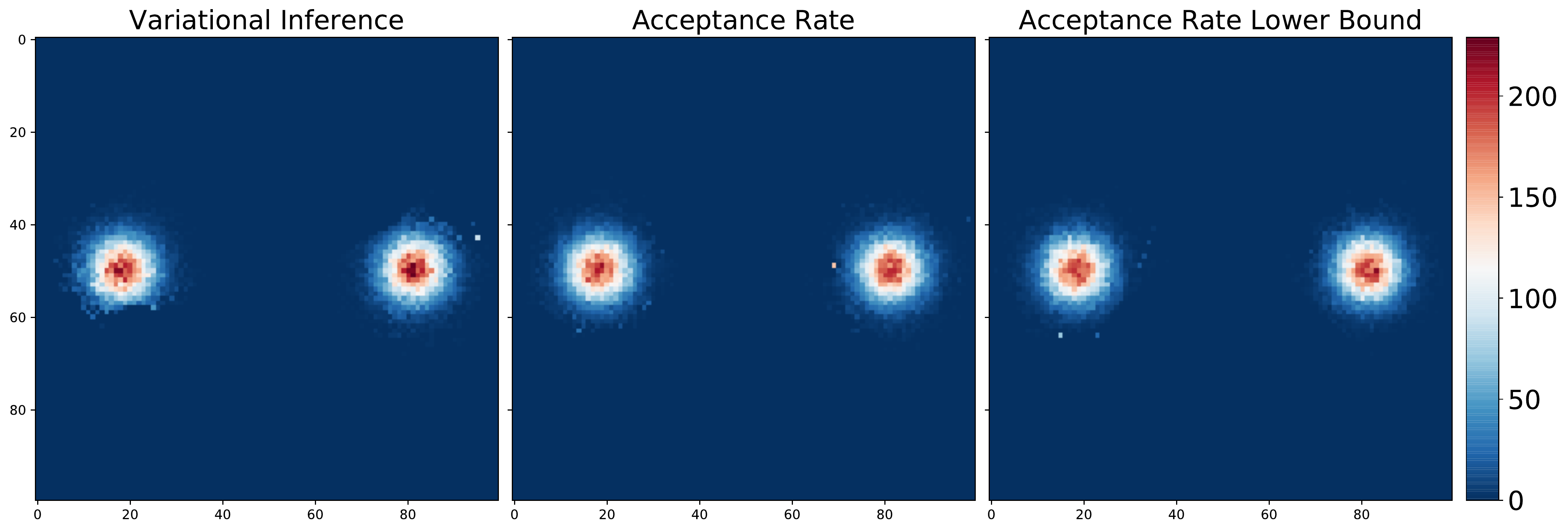}
    \\
    \includegraphics[width=0.49\textwidth]{pics/mog6_hist2d.pdf}
    \includegraphics[width=0.49\textwidth]{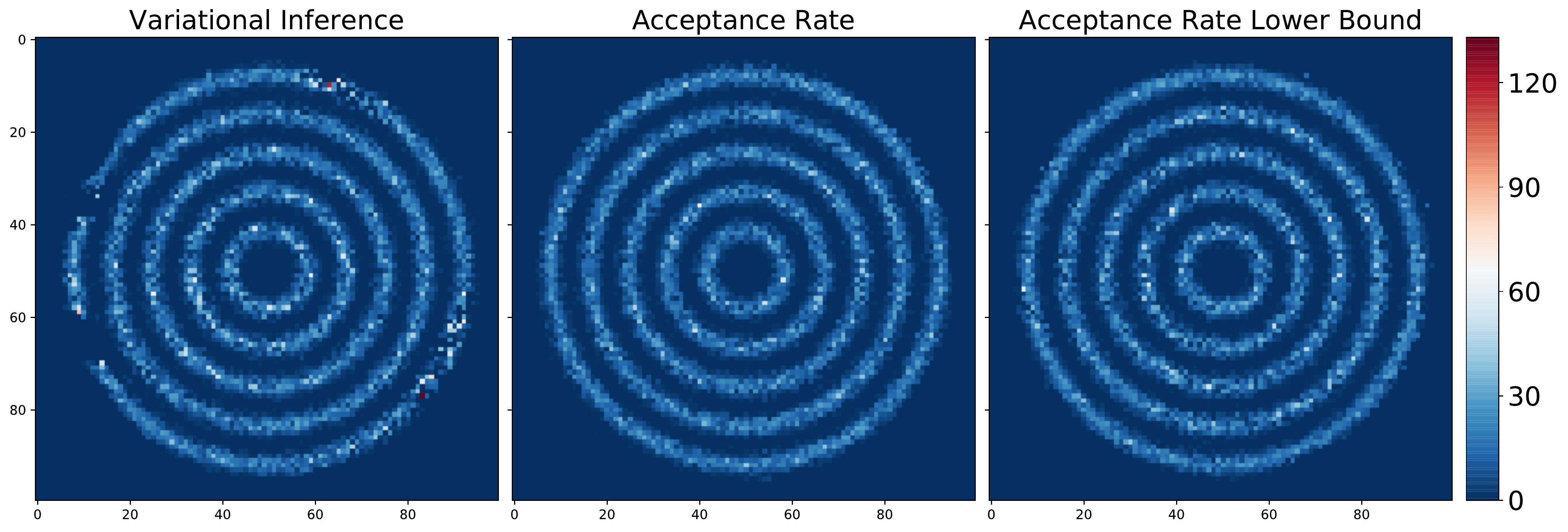}
    \caption{2d histrograms of samples from the MH algorithm with different proposals.
    For each distribution from left to right proposals are learned by:  variational inference, the acceptance rate maximization, the acceptance rate lower bound maximization.}
    \label{fig:hist2d}
\end{figure}

In this section we provide levelplots of learned proposals densities (see Fig. \ref{fig:density}). 
We also provide 2d histrograms of samples from the MH algorithm using the corresponding proposals (see Fig. \ref{fig:hist2d}).

\subsection{Mixing with a Random Walk}
\label{app:rw}

\begin{figure}[h]
    \centering
    \includegraphics[width=0.9\textwidth]{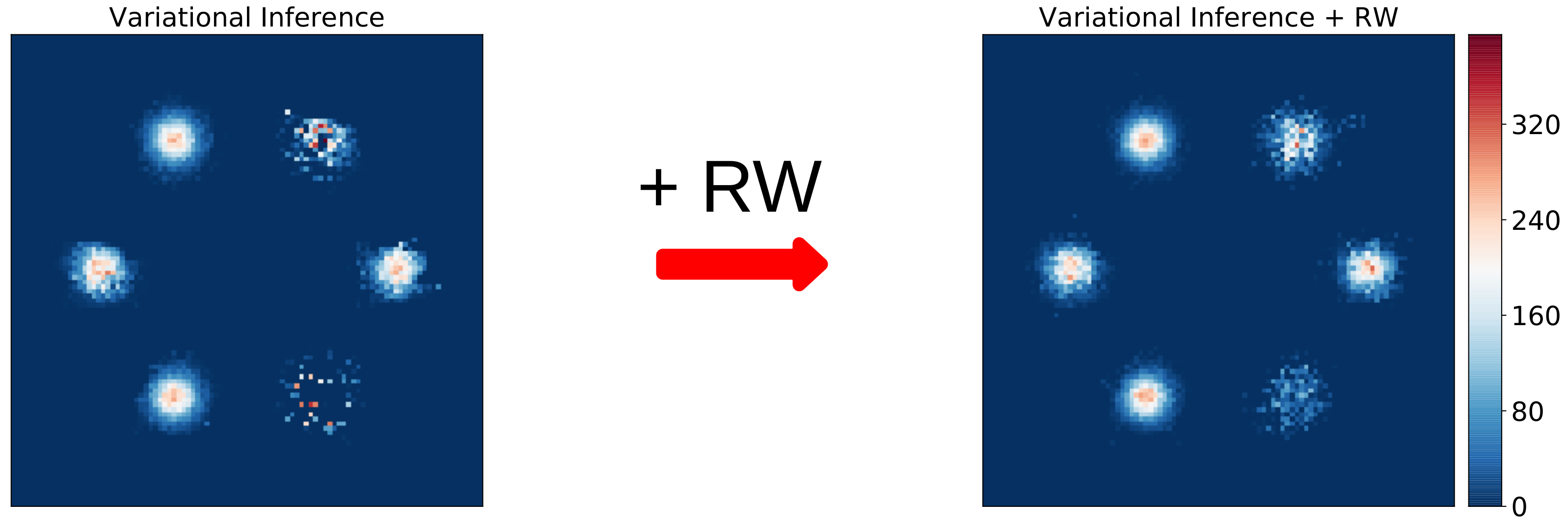}
    \caption{2d histrograms of samples from the MH algorithm for different proposals.
    On the left histogram we learn a proposal distribution with the variational inference.
    On the right histogram we mix the learned proposal with the random walk kernel as shown in Eq. \ref{eq:virw}.}
    \label{fig:virw}
\end{figure}

\begin{figure}[h]
    \centering
    \includegraphics[width=\textwidth]{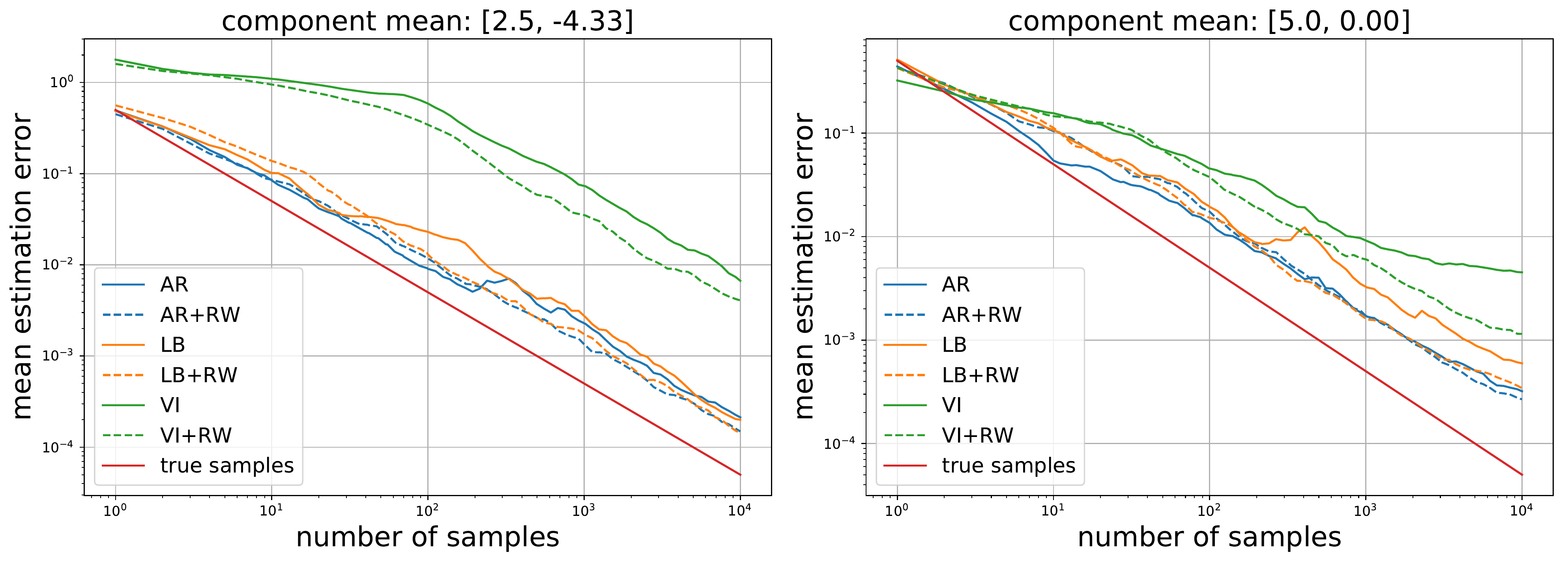}
    \caption{For \textbf{mog6} distribution we provide squared error of component's mean estimation depending on the number of samples.
    Here we run the MH algorithm until we have at least $10000$ samples around every component and then estimate their means.
    We average these plots across $100$ independent runs.
    Different plots correspond to different proposals: \textbf{AR} for the proposal learned by maximization of the acceptance rate; \textbf{LB} for the proposal learned by maximization of the acceptance rate lower bound; \textbf{VI} for the proposal learned by the variational inference.
    Suffix \textbf{+RW} states for mixing of a corresponding proposal with the random walk kernel.}
    \label{fig:comp_error}
\end{figure}

In this section, we follow \citep{de2001variational} and mix the proposal learned via the variational inference with a random walk kernel.
That is, given the target distribution density $p(x)$ we learn an independent proposal by optimization of the reversed KL-divergence:
\begin{equation}
    q^*(x) = \argmin_q \mathrm{KL}(q(x)||p(x)).
\end{equation}
Then we build a proposal for the Metropolis-Hastings algorithm by mixing the learned proposal $q^*(x)$ with the random walk kernel:
\begin{equation}
    q(x'\cond x) = \lambda q^*(x') + (1-\lambda) \Normal(x'\cond x, \sigma^2), \;\;\; \lambda \in [0,1].
    \label{eq:virw}
\end{equation}
According to \citep{de2001variational}, the main idea of such a mixture is that the variational approximation convergences quickly to the regions of a high target density and the random walk kernel describes neighbourhood of these regions.
In total agreement with this intuition, mixing with a random walk kernel does not improve the ESS, since ESS depends on statistics of the whole target distribution.
However, in Fig. \ref{fig:virw} we see that random walk improves mixing within individual components of the \textbf{mog6} distribution.
For numerical comparison, we estimate means of each component by samples from its neighbourhood and evaluate the squared error of estimation for different number of samples.
In Fig. \ref{fig:comp_error} we see that mixing with the random walk kernel stably improves error of the mean estimation for the proposal learned with the variational inference.
Moreover, we see that in some cases it improves mean estimation for proposals learned by the acceptance rate maximization.

\subsection{Bayesian Deep Learning experiments}
\label{app:bdl}

In density-based setting, we consider Bayesian inference problem for the weights of a neural network.
In our experiments we consider approximation of predictive distribution \eqref{eq:pred_dist} as our main goal.
To estimate the goodness of the approximation we measure negative log-likelihood and accuracy on the test set.

In subsection \ref{sec:db_algs} we show that lower bound on acceptance rate can be optimized more efficiently than acceptance rate due to the usage of minibatches.
But other questions arise.
\begin{enumerate}
    \item Does the proposed objective in \eqref{eq:db_inference} allow for better estimation of predictive distribution compared to the variational inference?
    \item Does the application of the MH correction to the learned proposal distribution allow for better estimation of the predictive distribution \eqref{eq:pred_dist} than estimation via raw samples from the proposal? 
\end{enumerate}

To answer these questions we consider reduced LeNet-5 architecture (see Appendix \ref{app:lenet}) for classification task on 20k images from MNIST dataset (for test data we use all of the MNIST test set).
Even after architecture reduction we still face a challenging task of learning a complex distribution in $8550$-dimensional space.
For the proposal distribution we use fully-factorized gaussian $q_\phi(\theta) = \prod_{j=1}^d\Normal(\theta_j \cond \mu_j,\sigma_j)$ and standard normal distribution for prior $p(\theta) = \prod_{j=1}^d\Normal(\theta_j \cond 0,1)$.

For variational inference, we train the model using different initialization and pick the model according to the best ELBO. 
For our procedure, we do the same and choose the model by the maximum value of the acceptance rate lower bound.
In Algorithm \ref{alg:db} we propose to sample from the posterior distribution using the independent MH and the current proposal.
In practice, better estimation of loss $\mathcal{L}(\phi)$ can be obtained by random-walk MH algorithm with currently learned proposal $q_\phi(\theta) = \Normal(\theta \cond \bm{\mu}, \bm{\sigma})$ as an initial state.
That is, we start with the mean $\bm{\mu}$ as an initial point, and then use random-walk proposal $q(\theta'\cond \theta) = \Normal(\theta' \cond \theta, \bm{\sigma})$ with the variances $\bm{\sigma}$ of current independent proposal.
This should be considered as a heuristic that improves the approximation of the loss function.
However, for test evaluation we use \textit{independent MH} with learned proposal.

The optimization of the acceptance rate lower bound results in the better estimation of predictive distribution than the variational inference (see Fig. \ref{fig:lenet5_perf}).
Optimization of acceptance rate for the same number of epochs results in nearly $30\%$ accuracy on the test set. 
That is why we do not report results for this procedure in Fig. \ref{fig:lenet5_perf}.

\begin{minipage}[h]{0.48\textwidth}
\centering
\includegraphics[width=\textwidth]{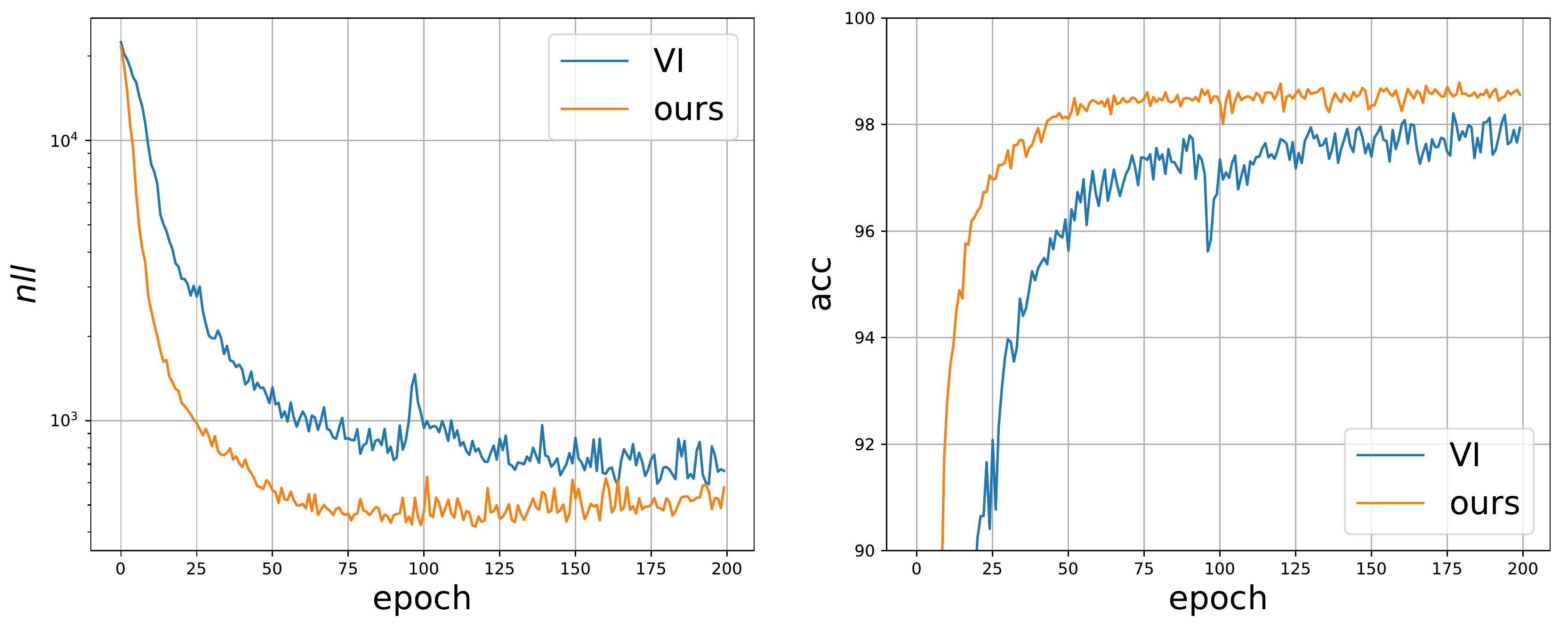}
\captionof{figure}{Negative log-likelihood (left) and accuracy (right) on test set of MNIST dataset for variational inference (blue lines) and the optimization of the acceptance rate lower bound (orange lines).
In both procedures we apply the independent MH algorithm to estimate the predictive distribution.}
\label{fig:lenet5_perf}
\end{minipage}\hfill
\begin{minipage}[h]{0.48\textwidth}
\centering
\includegraphics[width=\textwidth]{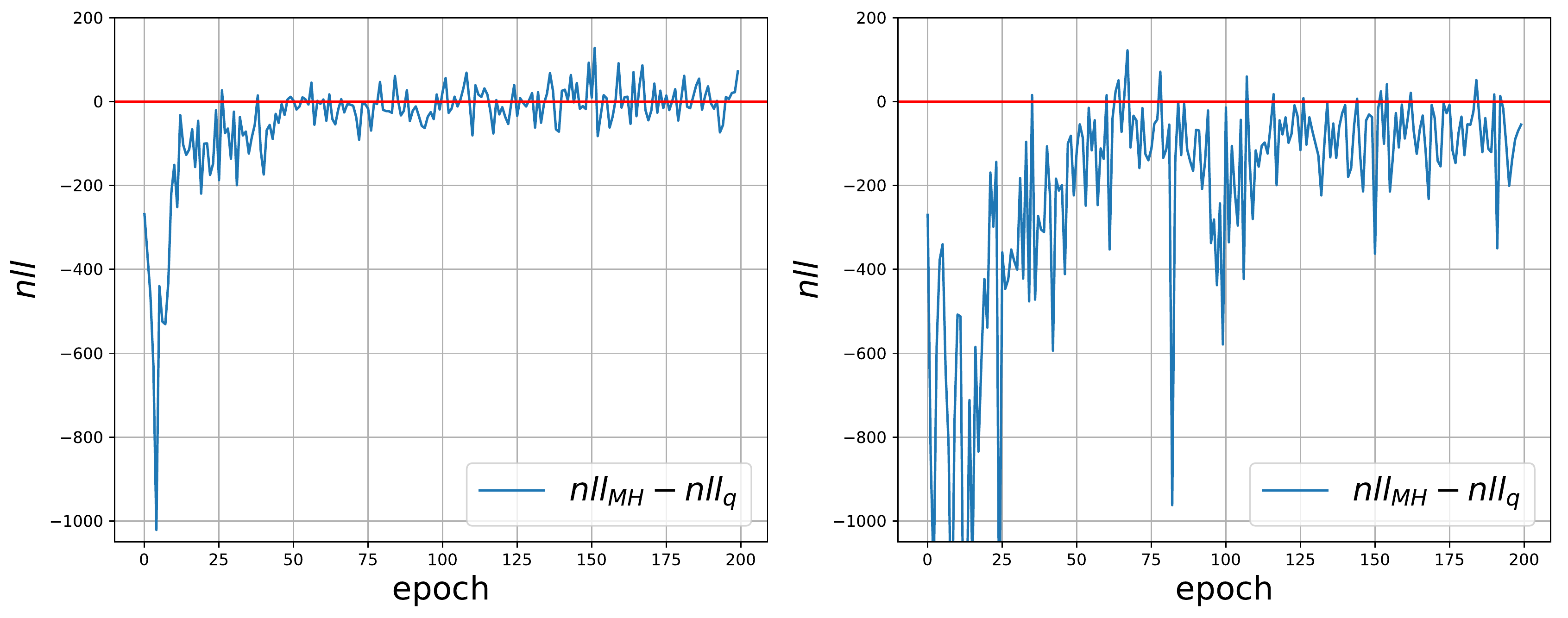}
\captionof{figure}{Test negative log-likelihood for two approximations of the predictive distribution based on samples: from proposal distribution $nll_{q}$ and after MH correction $nll_{MH}$.
    Left figure corresponds to the optimization of the acceptance rate lower bound, right figure corresponds to  the variational inference.}
\label{fig:ind_MH_vs_Q}
\end{minipage}

To answer the second question we estimate predictive distribution in two ways.
The first way is to perform $100$ accept/reject steps of the independent MH algorithm with the learned proposal $q_{\phi}(\theta)$ after each epoch, i.e. perform MH correction of the samples from the proposal. 
The second way is to take the same number of samples from $q_\phi(\theta)$ without MH correction.
For both estimations of predictive distribution, we evaluate negative log-likelihood on the test set and compare them.

The MH correction of the learned proposal improves the estimation of predictive distribution for the variational inference (right plot of Fig. \ref{fig:ind_MH_vs_Q}) but does not do so for the optimization of the acceptance rate lower bound (left plot of Fig. \ref{fig:ind_MH_vs_Q}).
This fact may be considered as an implicit evidence that our procedure learns the proposal distribution with higher acceptance rate.

\subsection{Architecture of the reduced LeNet-5}
\label{app:lenet}
\begin{verbatim}
class LeNet5(BayesNet):
    def __init__(self):
        super(LeNet5, self).__init__()
        self.num_classes = 10
        self.conv1 = layers.ConvFFG(1, 10, 5, padding=0)
        self.relu1 = nn.ReLU(True)
        self.pool1 = nn.MaxPool2d(2, padding=0)
        self.conv2 = layers.ConvFFG(10, 20, 5, padding=0)
        self.relu2 = nn.ReLU(True)
        self.pool2 = nn.MaxPool2d(2, padding=0)
        self.flatten = layers.ViewLayer([20*4*4])
        self.dense1 = layers.LinearFFG(20*4*4, 10)
        self.relu3 = nn.ReLU()
        self.dense2 = layers.LinearFFG(10, 10)
\end{verbatim}

\subsection{Markov proposal: MNIST experiments}
\label{app:mnist_gan}

In this section, we show proof of concept experiments on acceptance rate maximization for Markov proposal.
We use network architecture with bottleneck and dropout layers in "encoding" part to prevent collapsing to delta-function.

In the case of the Markov chain proposal, we show that the direct optimization of acceptance rate results in slow mixing --- most of the time the proposal generates samples from one of the modes (digits) and rarely switches to another mode.
When we perform the optimization of the lower bound the proposal switches between modes frequently (see Fig. \ref{fig:Markov_MH_filter}).
To show that the learned proposal distribution has the Markov property rather than being totally independent, we show samples from the proposal conditioned on two different points in the dataset (see Fig. \ref{fig:sb_onestep}).
Additionally, we demonstrate samples from the chain after $10000$ accepted images (see Fig. \ref{fig:thousand_markov}) and also samples from the chain that was initialized with noise (see Fig. \ref{fig:noise_markov}).

\begin{minipage}[h]{0.48\textwidth}
\centering
\includegraphics[width=0.49\textwidth]{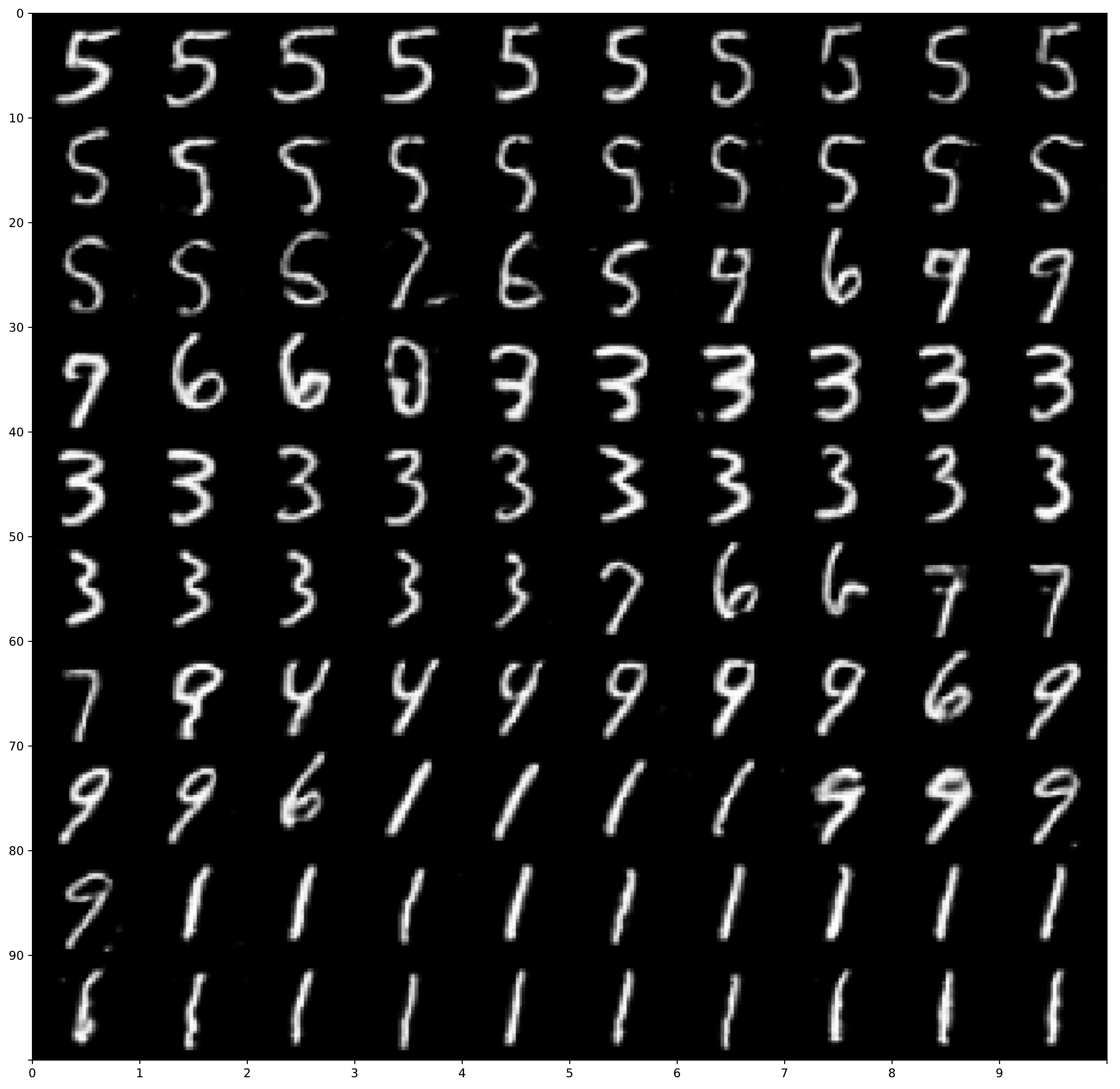}
\includegraphics[width=0.49\textwidth]{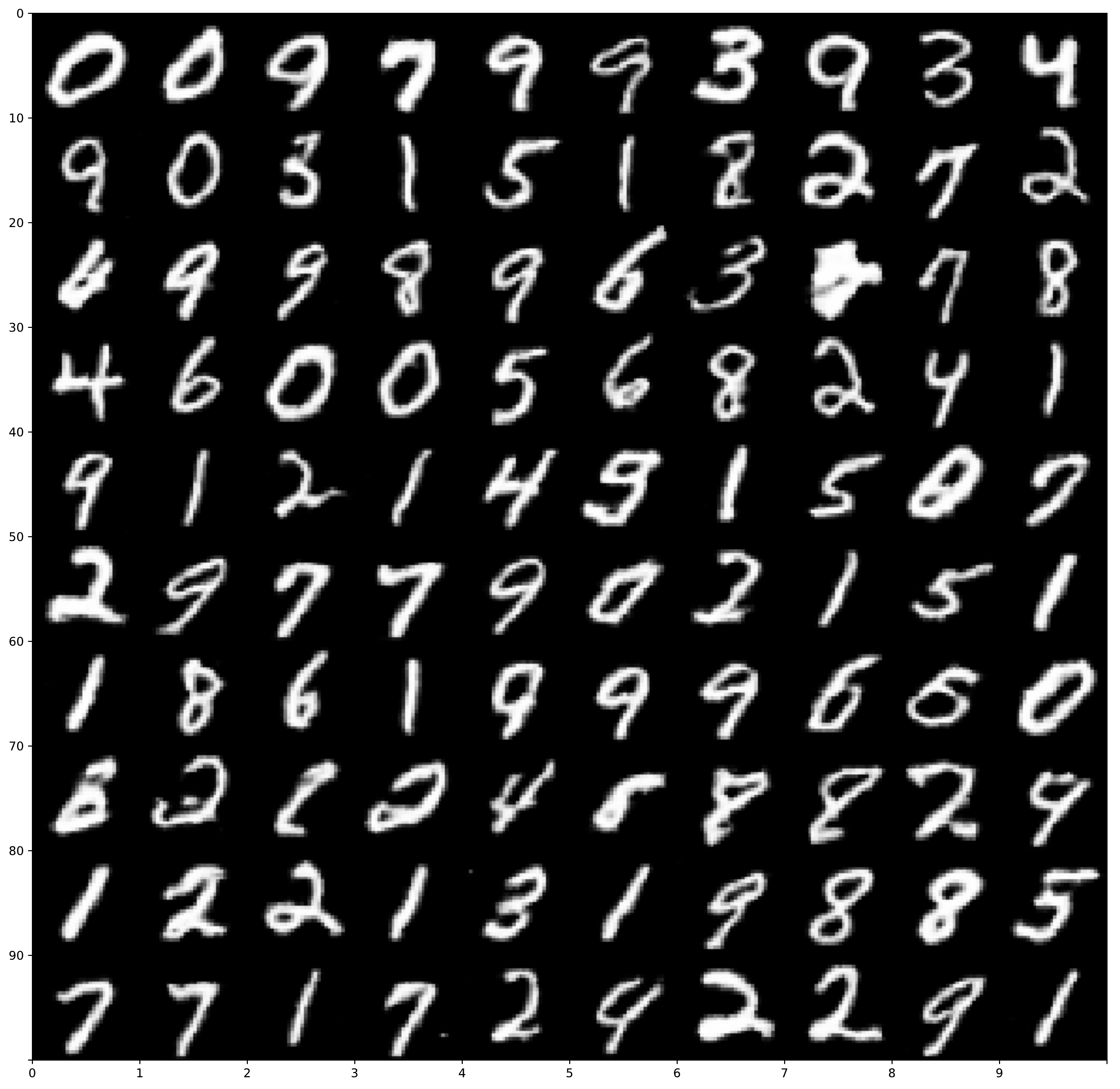}
\captionof{figure}{Samples from the chain obtained via the MH algorithm with the learned proposal and the learned discriminator for density ratio estimation.
The left figure corresponds to the direct optimization of the acceptance rate. 
The right figure -- to optimization of the lower bound on acceptance rate.
Samples in the chain are obtained one by one from left to right from top to bottom.}
\label{fig:Markov_MH_filter}
\end{minipage}\hfill
\begin{minipage}[h]{0.48\textwidth}
\centering
\includegraphics[width=0.49\textwidth]{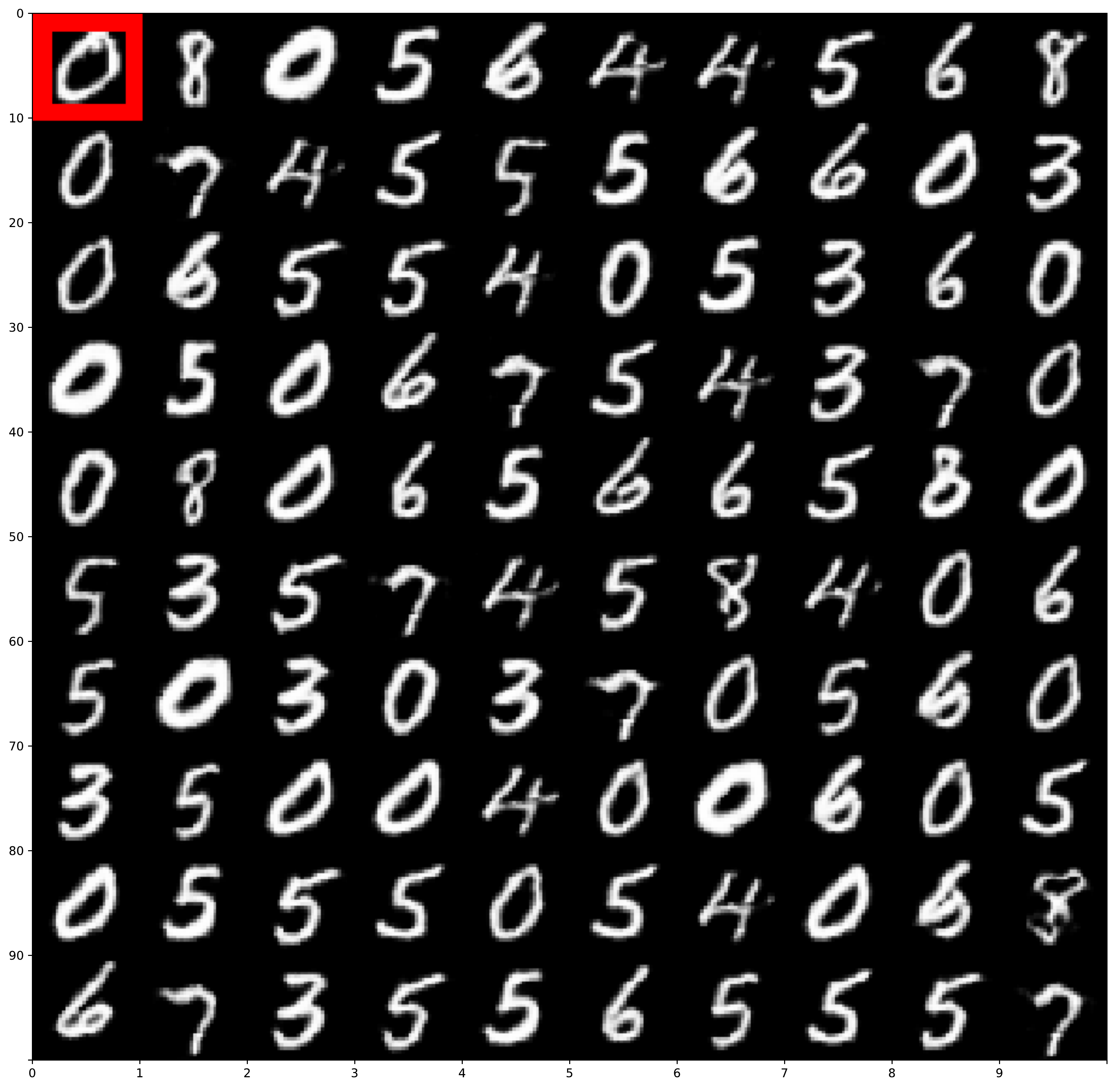}
\includegraphics[width=0.49\textwidth]{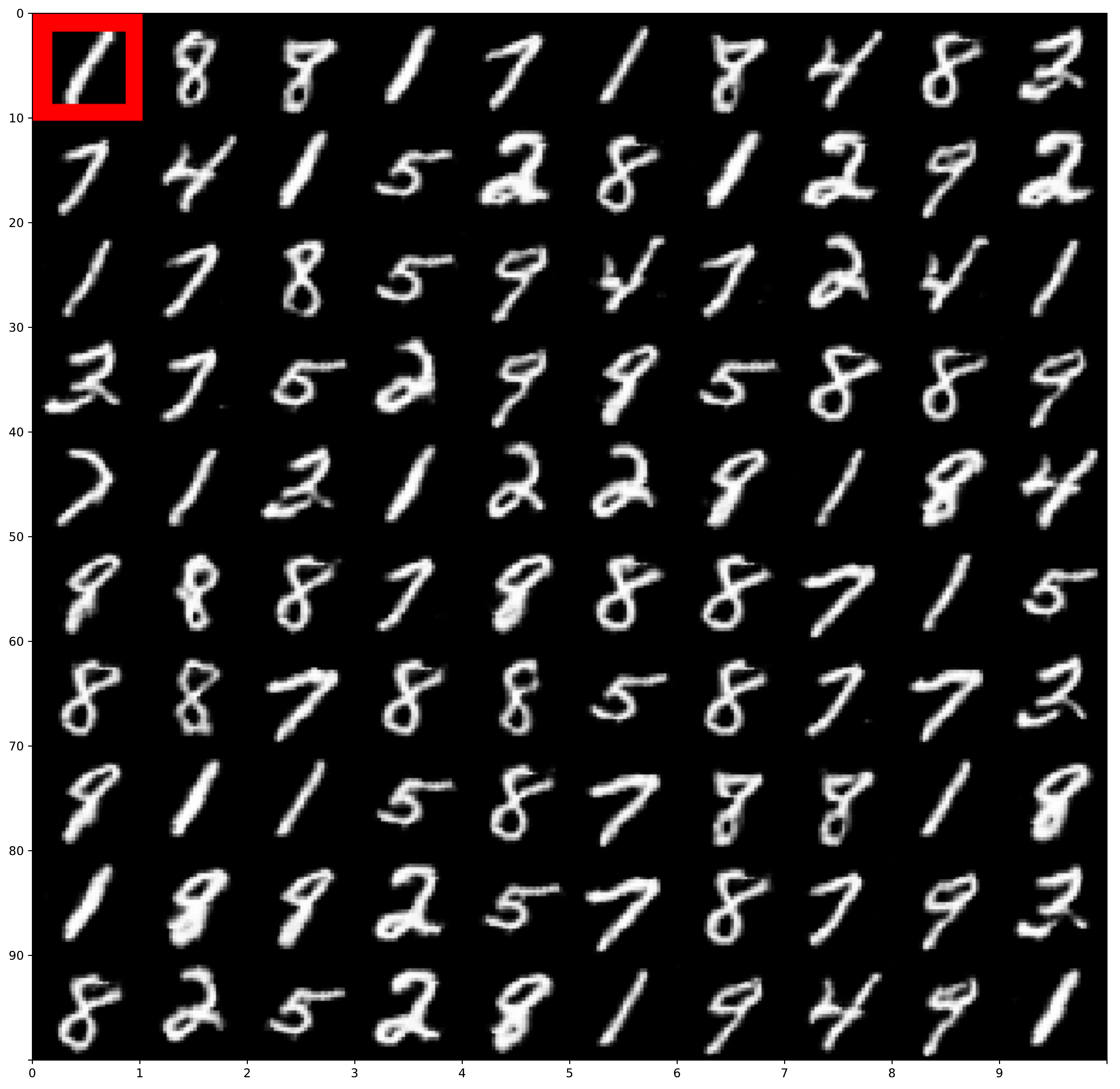}
\captionof{figure}{Samples from the proposal distribution and conditioned on the digit in the red box. 
The proposal was optimized according to the lower bound on the acceptance rate.
Note that we obtain different distributions of the samples because of conditioning of our proposal.}
\label{fig:sb_onestep}
\end{minipage}

\begin{minipage}[h]{0.48\textwidth}
\centering
\includegraphics[width=0.49\textwidth]{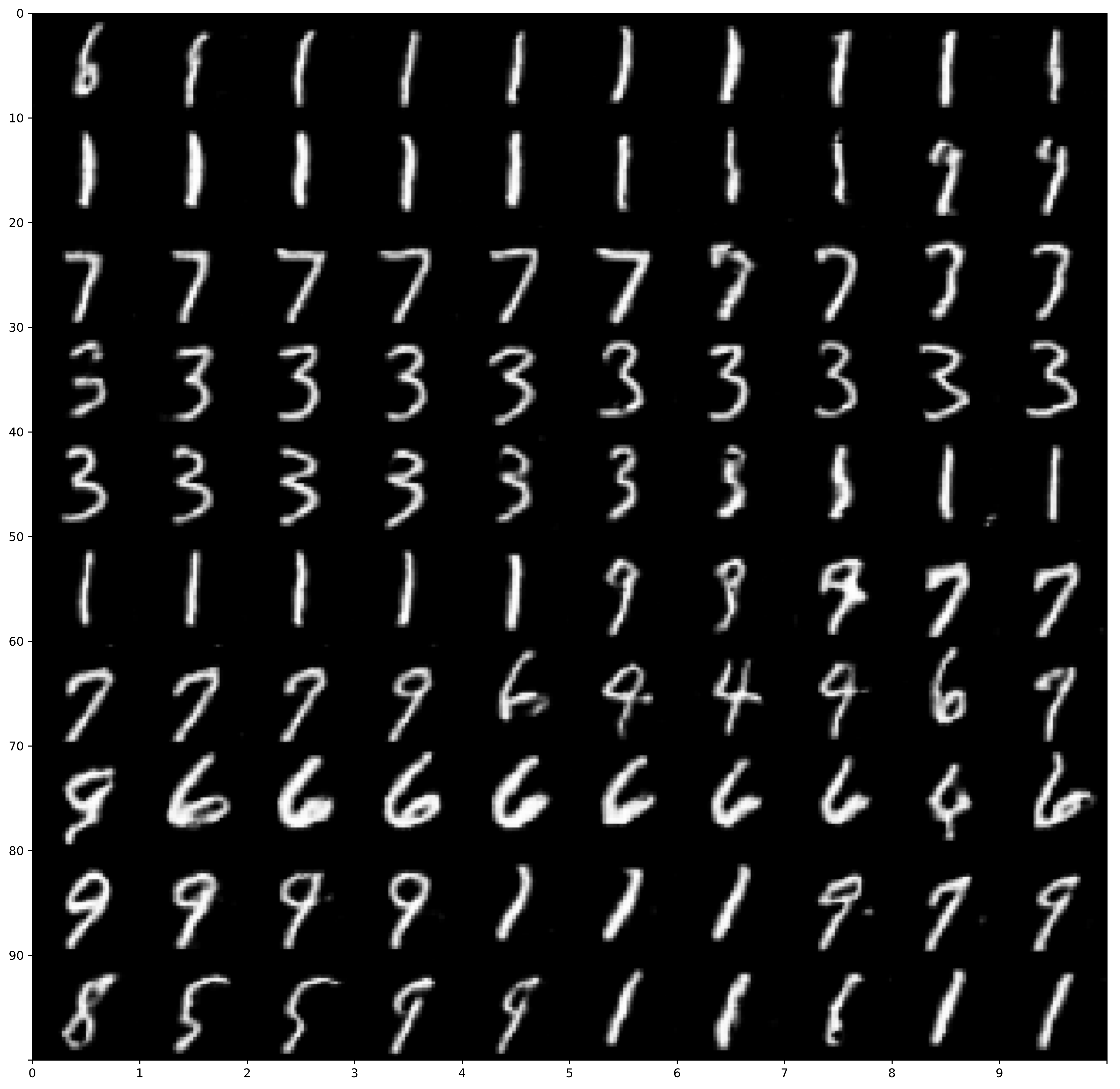}
\includegraphics[width=0.49\textwidth]{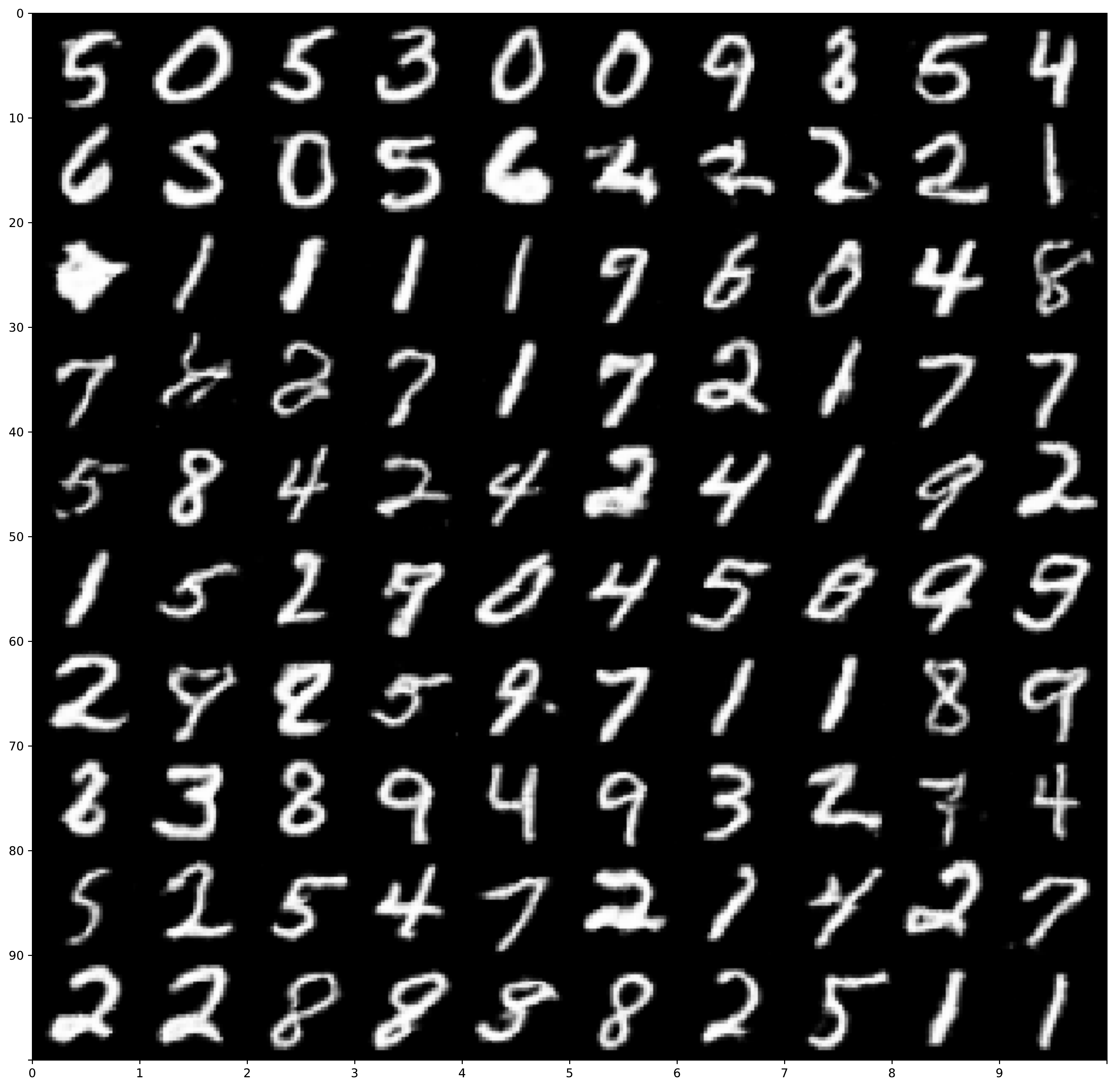}
\captionof{figure}{Samples from the chain after $10000$ accepted samples.
To obtain samples we use the MH algorithm with the learned proposal and the learned discriminator for density ratio estimation.
In the chain on the left figure we use proposal and discriminator that are learned during optimization of acceptance rate.
In the chain on the right figure we use proposal and discriminator that are learned during the optimization of the acceptance rate lower bound.
Samples in chain are obtained one by one from left to right from top to bottom.}
\label{fig:thousand_markov}
\end{minipage} \hfill
\begin{minipage}[h]{0.48\textwidth}
\centering
\includegraphics[width=0.49\textwidth]{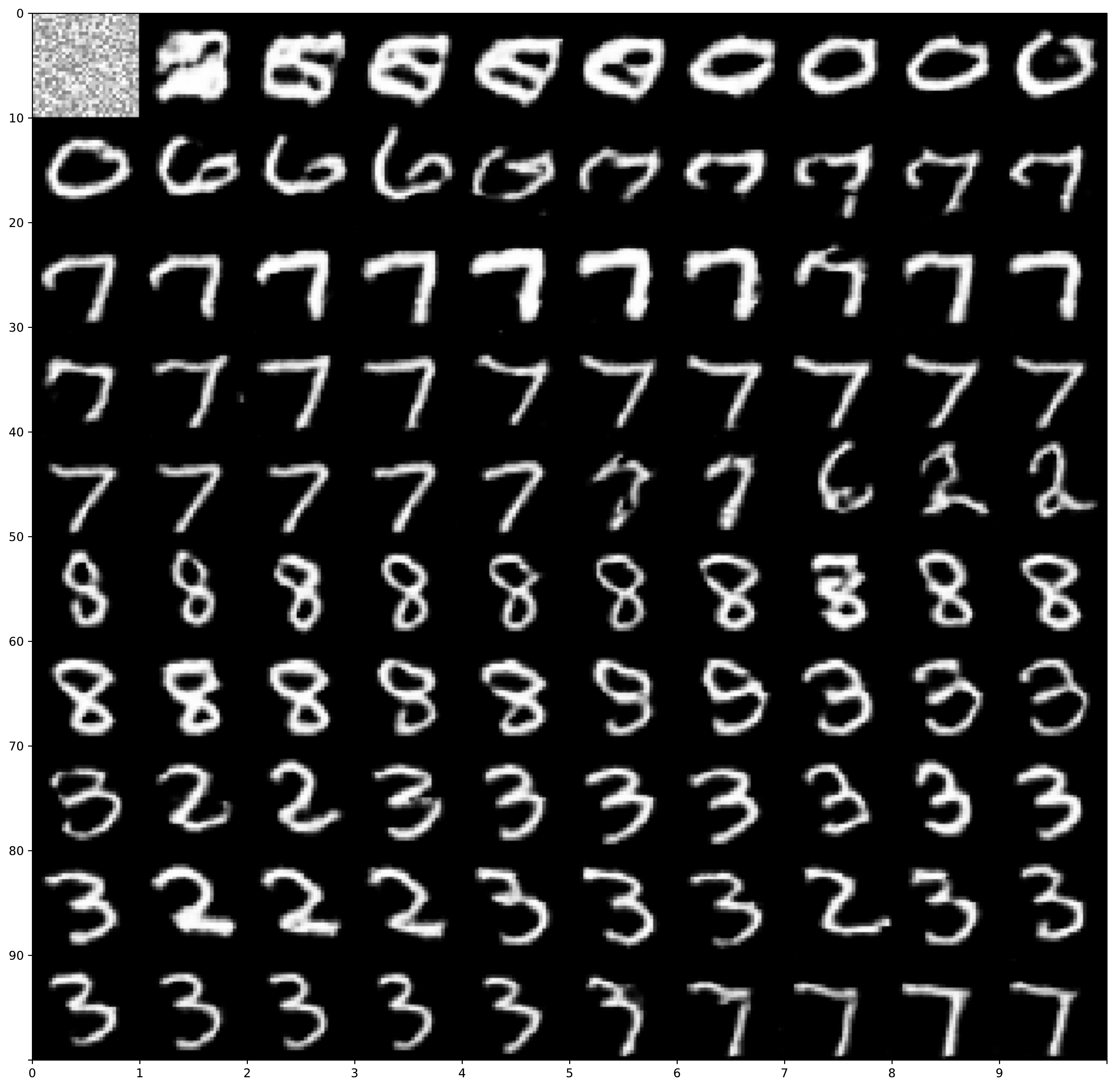}
\includegraphics[width=0.49\textwidth]{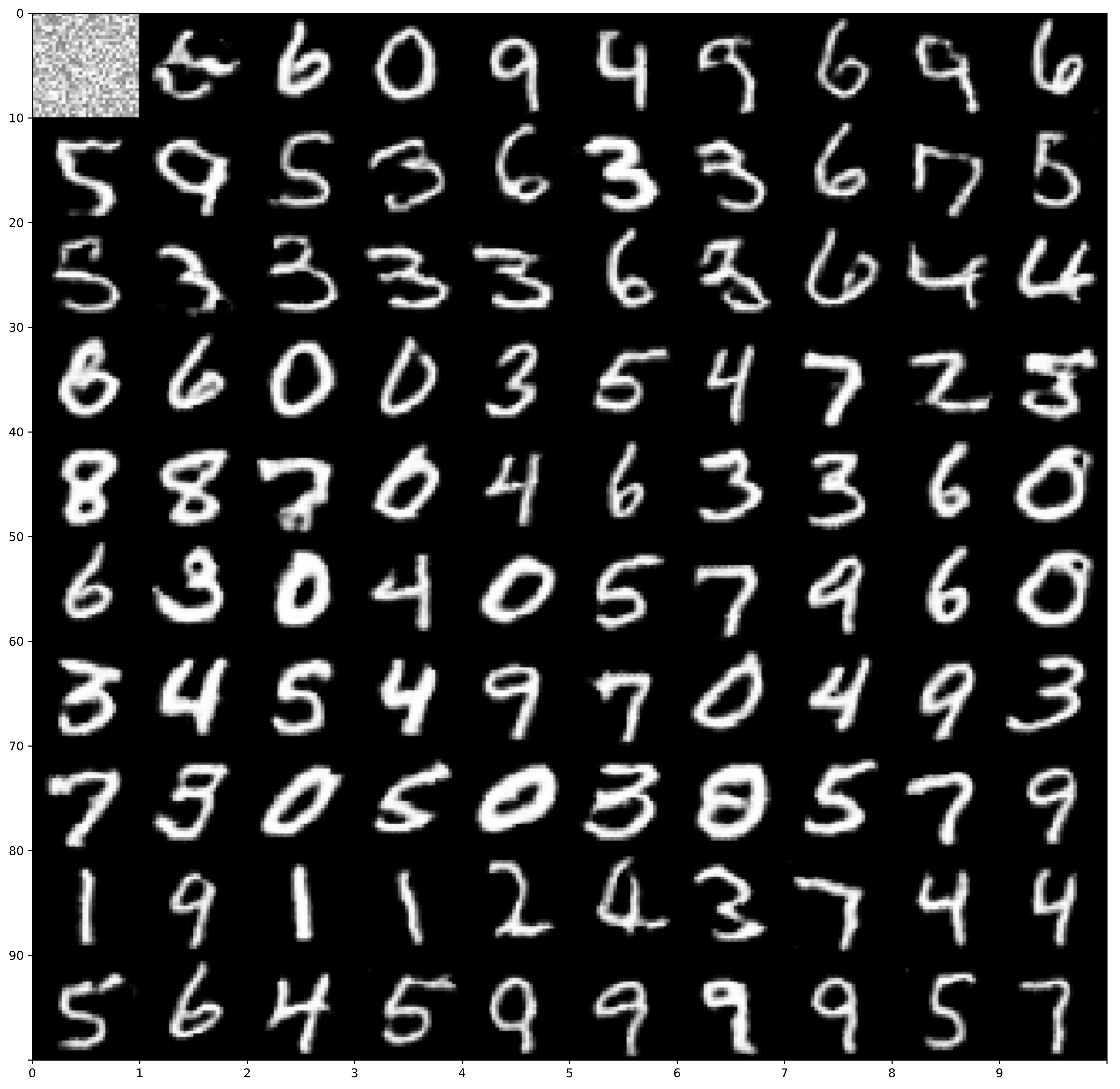}
\captionof{figure}{Samples from the chain initialized with noise.
To obtain samples we use the MH algorithm with the learned proposal and the learned discriminator for density ratio estimation.
In the chain on the left figure we use proposal and discriminator that are learned during optimization of acceptance rate.
In the chain on the right figure we use proposal and discriminator that are learned during the optimization of the acceptance rate lower bound.
Samples in the chain are obtained one by one from left to right from top to bottom starting with noise (first image in the figure).}
\label{fig:noise_markov}
\end{minipage}

\subsection{Architectures for Markov proposal}
\label{app:archs}

For Markov chain proposal distribution we use modified architecture of DCGAN.
\begin{verbatim}
class Generator(layers.ModuleWrapper):
    def __init__(self):
        super(Generator, self).__init__()

        self.d_conv1 = nn.Conv2d(1, 16, 5, stride=2, padding=2)
        self.d_lrelu1 = nn.LeakyReLU(0.2, inplace=True)
        self.d_do1 = nn.Dropout2d(0.5)
        self.d_conv2 = nn.Conv2d(16, 4, 5, stride=2, padding=2)
        self.d_in2 = nn.InstanceNorm2d(4, 0.8)
        self.d_lrelu2 = nn.LeakyReLU(0.2, inplace=True)
        self.d_do2 = nn.Dropout2d(0.5)

        self.b_view = layers.ViewLayer([4*8*8])
        self.b_fc = nn.Linear(4*8*8, 256)
        self.b_lrelu = nn.LeakyReLU(0.2, inplace=True)
        self.b_fc = nn.Linear(256, 128 * 8 * 8)
        self.b_do = layers.AdditiveNoise(0.5)

        self.e_unflatten = layers.ViewLayer([128, 8, 8])
        self.e_in1 = nn.InstanceNorm2d(128, 0.8)
        self.e_us1 = nn.ConvTranspose2d(128, 128, 2, 2)
        self.e_conv1 = nn.Conv2d(128, 128, 3, stride=1, padding=1)
        self.e_in2 = nn.InstanceNorm2d(128, 0.8)
        self.e_lrelu1 = nn.LeakyReLU(0.2, inplace=True)
        self.e_us2 = nn.ConvTranspose2d(128, 128, 2, 2)
        self.e_conv2 = nn.Conv2d(128, 64, 3, stride=1, padding=1)
        self.e_in3 = nn.InstanceNorm2d(64, 0.8)
        self.e_lrelu2 = nn.LeakyReLU(0.2, inplace=True)
        self.e_conv3 = nn.Conv2d(64, 1, 3, stride=1, padding=1)
        self.e_tanh = nn.Tanh()
\end{verbatim}
For density ratio we use discriminator of the following architecture.
\begin{verbatim}
class Discriminator(nn.Module):
    def __init__(self):
        super(Discriminator, self).__init__()
        self.conv1 = nn.Conv2d(2, 16, 3, 2, 1)
        self.lrelu1 = nn.LeakyReLU(0.2, inplace=True)
        self.conv2 = nn.Conv2d(16, 32, 3, 2, 1)
        self.lrelu2 = nn.LeakyReLU(0.2, inplace=True)
        self.in2 = nn.InstanceNorm2d(32, 0.8)
        self.conv3 = nn.Conv2d(32, 64, 3, 2, 1)
        self.lrelu3 = nn.LeakyReLU(0.2, inplace=True)
        self.in3 = nn.InstanceNorm2d(64, 0.8)
        self.conv4 = nn.Conv2d(64, 128, 3, 2, 1)
        self.lrelu4 = nn.LeakyReLU(0.2, inplace=True)
        self.in4 = nn.InstanceNorm2d(128, 0.8)
        self.flatten = layers.ViewLayer([128*2*2])
        self.fc = nn.Linear(128*2*2, 1)

    def forward(self, x, y):
        xy = torch.cat([x, y], dim=1)
        for module in self.children():
            xy = module(xy)
        yx = torch.cat([y, x], dim=1)
        for module in self.children():
            yx = module(yx)
        return F.softmax(torch.cat([xy, yx], dim=1), dim=1)
\end{verbatim}

\end{document}